\documentclass{article} 
\usepackage{iclr2026_conference,times}

\usepackage{amsmath,amsfonts,bm}

\def\eqref#1{equation~\ref{#1}}

\def\1{\bm{1}}

\DeclareMathAlphabet{\mathsfit}{\encodingdefault}{\sfdefault}{m}{sl}
\SetMathAlphabet{\mathsfit}{bold}{\encodingdefault}{\sfdefault}{bx}{n}

\usepackage{hyperref}
\usepackage{url}
\usepackage[utf8]{inputenc} 
\usepackage[T1]{fontenc}    
\usepackage{booktabs}       
\usepackage{amsfonts}       
\usepackage{nicefrac}       
\usepackage{microtype}      
\usepackage{lipsum}
\usepackage{fancyhdr}       
\usepackage{graphicx}       
\usepackage{pifont}
\usepackage{makecell}
\usepackage{wrapfig} 
\usepackage[table]{xcolor}
\usepackage{amsmath}
\usepackage{multirow}
\graphicspath{{media/}}     
\definecolor{darkgreen}{RGB}{0,100,0}
\definecolor{pink}{RGB}{255,105,180}

\title{Agent-ScanKit: Unraveling Memory and Reasoning of Multimodal Agents via Sensitivity Perturbations}

\iclrfinalcopy
\author{Pengzhou Cheng\textsuperscript{1}, 
  Lingzhong Dong\textsuperscript{1}, 
  Zeng Wu\textsuperscript{1}, 
  Zongru Wu\textsuperscript{1}, 
  Xiangru Tang\textsuperscript{2}, 
  Chengwei Qin\textsuperscript{3} \\
  \textbf{Zhuosheng Zhang}\textsuperscript{1}\thanks{Correspondence to Zhuosheng Zhang <zhangzs@sjtu.edu.cn> and Gongshen Liu <lgshen@sjtu.edu.cn> }, 
  \textbf{Gongshen Liu}\textsuperscript{1*} \\
  \textsuperscript{1}Shanghai Jiao Tong University \quad
  \textsuperscript{2}Yale University \quad \\
  \textsuperscript{3}The Hong Kong University of Science and Technology (Guangzhou) \\
}

\begin{document}

\maketitle

\begin{abstract}
Although numerous strategies have recently been proposed to enhance the autonomous interaction capabilities of multimodal agents in graphical user interface (GUI), their reliability remains limited when faced with complex or out-of-domain tasks. This raises a fundamental question: Are existing multimodal agents reasoning spuriously?  In this paper, we propose \textbf{Agent-ScanKit}, a systematic probing framework to unravel the memory and reasoning capabilities of multimodal agents under controlled perturbations. Specifically, we introduce three orthogonal probing paradigms: visual-guided, text-guided, and structure-guided, each designed to quantify the contributions of memorization and reasoning without requiring access to model internals. In five publicly available GUI benchmarks involving 18 multimodal agents, the results demonstrate that mechanical memorization often outweighs systematic reasoning. Most of the models function predominantly as retrievers of training-aligned knowledge, exhibiting limited generalization. Our findings underscore the necessity of robust reasoning modeling for multimodal agents in real-world scenarios, offering valuable insights toward the development of reliable multimodal agents. 
Our code is available at~\url{https://github.com/CTZhou-byte/Agent_ScanKit}.
\end{abstract}

\section{Introduction}
\begin{figure}[t]
    \centering
    \includegraphics[width=1\linewidth]{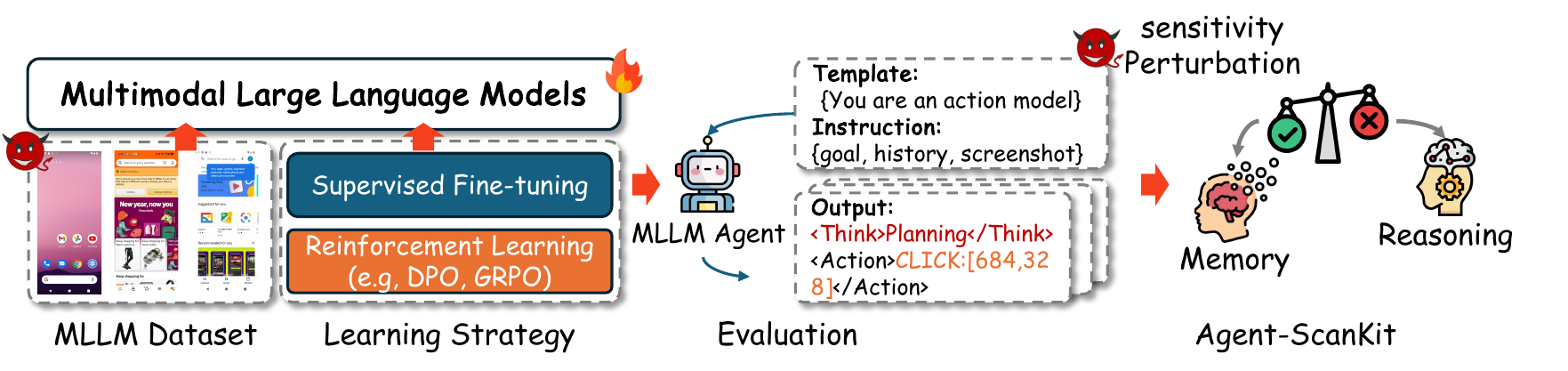}
    \vspace{-0.4cm}
    \caption{Pipeline of existing multimodal agents for GUI tasks. However, their poor reliability may stem from reliance on memorization rather than genuine reasoning.}
    \vspace{-0.5cm}
    \label{fig1}
\end{figure}

With recent advances in multimodal large language models (MLLMs)~\citep{hurst2024gpt, qwen2.5-VL, shen2025vlm}, building multimodal agents has become more straightforward and generalizable, particularly in graphical user interfaces (GUIs). These agents promise broad task automation on mobile and desktop devices~\citep{wang2024gui, zhang2024large}. Compared to previous agents that relied on textual descriptions of the environment, such as HTML or accessibility trees, MLLM-based GUI agents predict the subsequent action based on a specific goal with only environmental perception (e.g., screen)~\citep{ma2024caution}. As shown in Figure~\ref{fig1}, recent work advances grounding~\citep{wu2024atlas, qin2025ui, zhou2025gui}, planning~\citep{zhang2024dynamic, wu2025vsc}, reflection~\citep{lu2025ui, luo2025gui, liu2025infigui}, and adaptation~\citep{bai2024digirl, wang2024distrl} through continue pretraining (CPT), supervised fine-tuning (SFT), and reinforcement learning (RL). Notable RL variants include Direct Preference Optimization (DPO)~\citep{rafailov2023direct} and Group Relative Policy Optimization (GRPO)~\citep{shao2024deepseekmath}.

However, existing open-source multimodal agents still exhibit poor reliability when faced with complex or out-of-domain (OOD) GUI tasks~\citep{wu2025gem,liu2025verigui,guo2025atomic}. Related studies further suggest that the so-called ``reasoning'' ability of LLMs often reduces to sophisticated pattern matching~\citep{mirzadehgsm} or even rote memorization of training data~\citep{carlini2021extracting, hartmann2023sok}. 
Therefore, we conduct a systematic analysis and identify three core contributors to unreliability.

First, the inherently unbounded nature of visual and textual spaces results in potential visual and textual-oriented \textit{ memory biases}, which decrease the accuracy of the prediction and directly undermine the success rates of the tasks. Second, prior research has focused primarily on learning within these two spaces, while overlooking the optimization of state and reflection action, thus introducing varying degrees of action-based \textit{memory shortcuts}. Third, \textit{domain sensitivity} further limits the agent’s ability to generalize across tasks and environments.
Although many models report stepwise accuracy (SR) above 80\% and task success rates above 40\% on benchmarks such as AITZ~\citep{zhang2024android} and AndroidControl~\citep{li2024effects} (Table~\ref{tab_a1}), our findings reveal a significant performance drop when these models are evaluated on long-horizon tasks or cross-platform scenarios (Table~\ref{tab_a2}). These observations motivate a key research question: \textit{Memory or reasoning: what drives multimodal agents?}

In this paper, we propose \textbf{Agent-ScanKit}, a systematic probing toolkit to unravel memory and reasoning capabilities in multimodal agents by sensitivity perturbation. Specifically, we introduce three orthogonal probing paradigms: 

(i) In the \textit{visual-guided level}, we use object masking and editing to test whether grounding relies on memorization, while a zooming strategy quantifies reasoning under local visual changes.

(ii) In the \textit{text-guided level}, we adopt atomic instruction masking in token-level and substitution in sentence-level, aiming to probe memory and reasoning in textual modal.

(iii) In the \textit{structure-level}, we probe that specific status and reflection actions are shortcuts memory, or reasoning caused by reflection. 

Each was designed to isolate and quantify the contributions of memorization and reasoning without requiring access to the model's internals. In five publicly available GUI benchmarks involving 18 agents, the results reveal that existing multimodal agents exhibit over-memorization in three probing strategies. Concretely, these agents tend to construct complex, brittle mappings between inputs and outputs, acting more as retrievers of training-aligned knowledge than as genuine reasoners. Furthermore, RL-based methods combined with the chain-of-thought (CoT) mechanism have some reasoning capabilities on language modal-side, enabling the competence extrapolation and environmental adaptability. These findings clearly define genuine reasoning mechanisms within multimodal agents for building more reliable and general-purpose AI assistants. Our contributions can be summarized as follows:

(i) We conduct a comprehensive evaluation of 18 open-source GUI agents on 5 benchmarks, revealing two central challenges: the infinite predictive space and finite generalization

(ii) We present Agent-ScanKit, a systematic probing toolkit, which provides a unified analysis across visual, textual, and structural dimensions through sensitivity perturbations, enabling quantitative assessment of memory and reasoning in multimodal agents.

(iii) We show that existing multimodal agents often display spurious reasoning behaviors driven by over-memorization. Although RL and CoT-augmented strategies have facilitated progress in GUI tasks, substantial room for improvement remains.

\section{Related Works}
This section reviews two lines of research that from the basis of this work: (i) multimodal agents for GUI interaction, and (ii) internal mechanisms for memory and reasoning.
\subsection{MLLM-based GUI Agents}
The rise of MLLMs~\citep{qwen2.5-VL, zhang2024large} has a significant shift in GUI automation, moving beyond rigid script- or rule-based systems~\citep{hellmann2011rule, steven2000jrapture}. By perceiving UI states (e.g., screenshots) and performing atomic actions like clicks and typing, multimodal agents enable more flexible, human-like interactions across platforms, such as Desktop~\citep{niu2024screenagent, zhang2024ufo, wu2024copilot}, Web~\citep{gurreal, zheng2024gpt, ma2023laser, shen2025thinking}, and Mobile~\citep{zhang2024you, zhang2025appagent}. This paper investigates the mechanism of memory and reasoning in multimodal agents in GUI tasks. Following SPA-Bench~\citep{chen2024spa} and RiOS-World~\citep{yang2025riosworld}, existing GUI agents can be categorized into two paradigms: agentic workflows and agent-as-a-model. 

The former is framework-based that adopts prompt learning on a proprietary model and leverages the power of MLLMs (e.g., GPT-4o and Claude 3.5 Sonnet) to build environment perception~\citep{zhang2025appagent, li2024appagent}, task planning~\citep{guo2025atomic}, decision reflection~\citep{rawles2024androidworld, liu2025infiguiagent}, memory persistence~\citep{dai2025advancing, jiang2025appagentx}, and multi-agent collaboration~\citep{wang2024mobile, wang2025mobile, wang2025mobile, zhang2024ufo, khaokaew2024maple}. However, practitioners have raised concerns about privacy leakage, the cost of API usage, and latency during inference on real-world devices. In addition, task performance is generally poorer compared to the latter. In contrast, the agent-as-a-model centers on building native agent models. By customizing MLLMs through CPT, SFT, and RL for agentic tasks, workflow knowledge is embedded directly into the model itself. This enables capabilities such as grounding enhancements~\citep{wu2024atlas, qin2025ui, zhou2025gui,zhang2025agentcpmgui, wu2025smoothing}, planning~\citep{ma2024coco, zhang2024dynamic, wu2025vsc}, reflection~\citep{zhang2024android, luo2025gui, lu2025ui, wanyan2025look, wu2025gui}, environmental adaptation~\citep{bai2024digirl, wang2024distrl, xie2025gui}, experience replay~\citep{liu2025learnact, zhang2025tongui} and reliability~\citep{ma2024caution, cheng2025kairos, cheng2025navi, wu2025verios}. Nevertheless, these models struggle on complex or OOD tasks, motivating us to quantify their memory and reasoning capabilities for deeper insight into their execution mechanisms.

\subsection{Memory vs. Reasoning} Recently, two perspectives on the execution mechanism of LLMs have emerged: reasoning vs. memory. The former has been demonstrated in tasks such as mathematics~\citep{wang2025reinforcement, luo2025ursa} and QA~\citep{chen2025sft, guo2025deepseek}, where LLMs appear to provide correct answers by CoT. However,  several studies have shown that the purported ``reasoning'' ability of LLMs is largely attributable to sophisticated pattern matching~\citep{mirzadehgsm,carlini2021extracting, hartmann2023sok}. They also investigated the formation and contribution of memories~\citep{speicher2024understanding, dankers2024generalisation}, and demonstrated that such mechanisms are effective primarily on simple tasks~\citep{li2024understanding, jin2025exploring}. Further studies highlight the importance of detecting and disentangling LLM memorization~\citep{djire2025memorization, jin2024disentangling}, as well as exploring how such mechanisms can be systematically measured~\citep{schwarzschild2024rethinking}. Thus, given the context of poor reliability of multimodal agents in GUI tasks, the quantification of memory and reasoning capabilities becomes critical.

\section{Challenge of Multimodal Agents}\label{sec3}
In this section, we first formalize the multimodal agent task execution process in GUI tasks. Then, we conduct a comprehensive evaluation of 18 agents on five benchmarks. For completeness, we report detailed results in the Appendix~\ref{B.2}.

\subsection{Problem Statement}
\noindent\textbf{MLLM-based GUI Agents.} Following prior works~\citep{wu2025vsc, wang2025spa}, we formalize GUI agentic tasks as a goal-driven partially observable Markov decision process (POMDP), defined by the tuple $\mathcal{M}=(G, \mathcal{S}, \mathcal{A}, \mathcal{T}, \mathcal{R}, \mathcal{H})$. Here, $G$ denotes the goal space, $\mathcal{S}$ the perceptual state space (e.g., screenshots and supplementary data~\citep{cheng2025kairos}),
$\mathcal{A}$ the action space (Appendix~\ref{action_space}), $\mathcal{T}: \mathcal{S} \times \mathcal{A} \rightarrow \mathcal{S}$ the transition function, $\mathcal{R}: \mathcal{S} \times \mathcal{A} \rightarrow \mathbb{R}$ the bounded reward function, typically positive upon task completion, and $\mathcal{H}$ the maximum action steps for a goal. 

Given a user goal $g \in G$, the agent observes the environment state $s_t$ at time $t$ and predicts an action $a_t \in \mathcal{A}$ through a structured reasoning process. This reasoning process may involve a CoT $r_t$, which allows the agent to interpret its observations and refine its decisions step by step, as seen in R1-like agents such as AgentCPM-GUI~\citep{zhang2025agentcpmgui} and GUI-Owl~\citep{ye2025mobileagentv3foundamentalagentsgui}. Executing $a_t$ yields the next state $s_{t+1}$, and the trajectory $(s_{1:n}, r_{1:n}, a_{1:n})$ constitutes an episode associated with $g$, formalized as $E=(g, \{s_t, r_t, a_t\}^n_{t=1})$. 

As just discussed, the development  of multimodal agents for GUI tasks is generally unified under a three-stage training framework~\citep{tang2025magicgui}, comprising perception enhancement through CPT, behavioral imitation via SFT, and generalization with RL, further reinforced by data and model scaling laws. Despite this progress, their capabilities remain constrained: agents often rely on visual-textual rule matching rather than understanding the operational logic of GUl. We refer to this phenomenon as memory-driven spurious reasoning. Meanwhile, we will also quantify whether genuine reasoning is present. To this end, we first conduct a comprehensive evaluation of existing agents and identify two key challenges: (i) infinite predictive space; (ii) finite generalization.

\subsection{Fails due to the Infinite predictive  Space}\label{3.2}
We divide the infinite predictive space into two categories: coordinate space and vocabulary space. This implies that for GUI agents to execute tasks autonomously, they must not only identify the correct action type but also predict accurately within these two unbounded spaces. Thus, we first investigate whether infinite predictive spaces contribute to the main inferior performance.
\begin{figure}[t]
    \centering
    \includegraphics[width=1\linewidth]{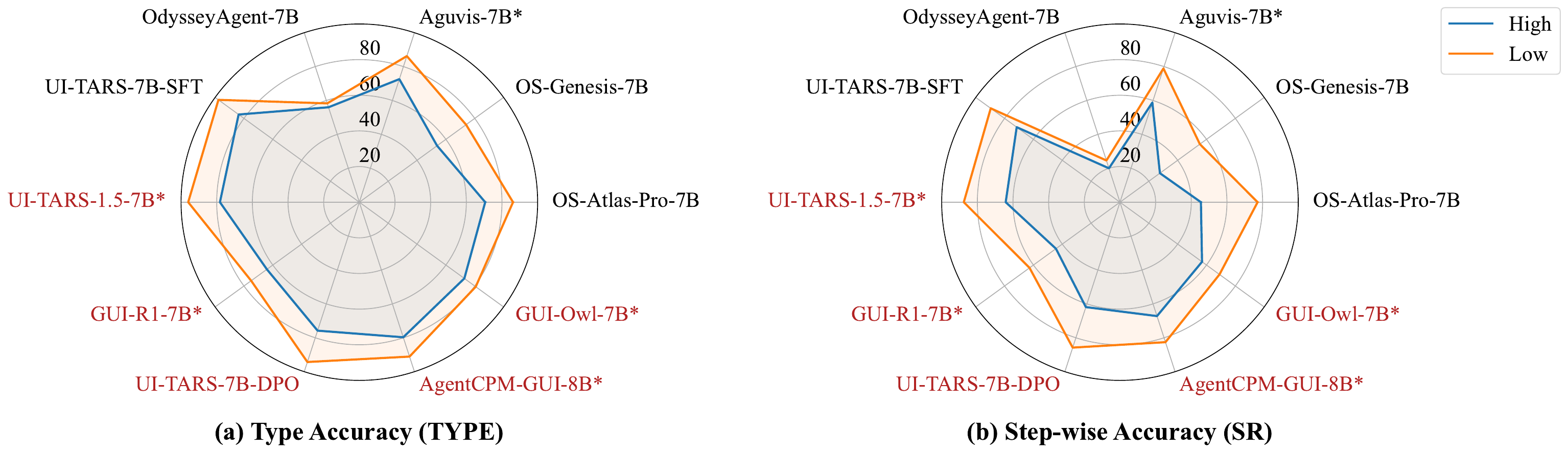}
    \caption{Comparative performance of 7$\sim$8B multimodal agents on two evaluation metrics in GUI tasks. RL-based models are highlighted in red, while reasoning-enabled models are marked with ``*''. Low-level provides atomic instructions based on queries, whereas high-level only offers the query.}
    \label{细粒度指标}
    \vspace{-0.4cm}
\end{figure}

To begin with, we present the performance for 10 multimodal agents with 7 to 8B scale under the AndroidControl benchmark~\citep{li2024effects}, as shown in Figure~\ref{细粒度指标}. Fine-grained action-type accuracy under the low-level setting is reported in the Appendix~\ref{Action Types}. Overall, existing muiltimodal agents perform relatively robustly on actions in coordinate-based (e.g., \textsc{Click}) and vocabulary-based (e.g., \textsc{Type}), yet with room for improvement, particularly within the vocabulary domain. However, the performance of reflection actions (e.g.\textsc{PressHome} and \textsc{PressBack}) and state actions (e.g., \textsc{Wait} and \textsc{Complete}) is poor. Importantly, SR accuracy declines sharply. We attribute this limitation to severe imbalances in training data and optimization strategies. Grounding data dominates the distribution, while augmentation strategies emphasize perceptual aspects. Thus, models overfit to coordinate space while under-exploiting vocabulary-based space. In addition, atomic instructions (low-level) significantly enhance agent performance, suggesting a potential text-guided reasoning mechanism. Therefore, within the infinite predictive space, detecting whether multimodal agents are relying on rote memorization or genuine reasoning is of paramount importance.

\subsection{Fails due to the finite Generalization}
We divide the finite generalization into task and environment categories. For GUI agents to execute tasks reliably, they also need to satisfy summarization and reasoning beyond their training data, thereby extending the generalization. We thus investigate whether finite generalization also underlie their inferior performance through task success rates.
\begin{figure}[t]
    \centering
    \includegraphics[width=1\linewidth]{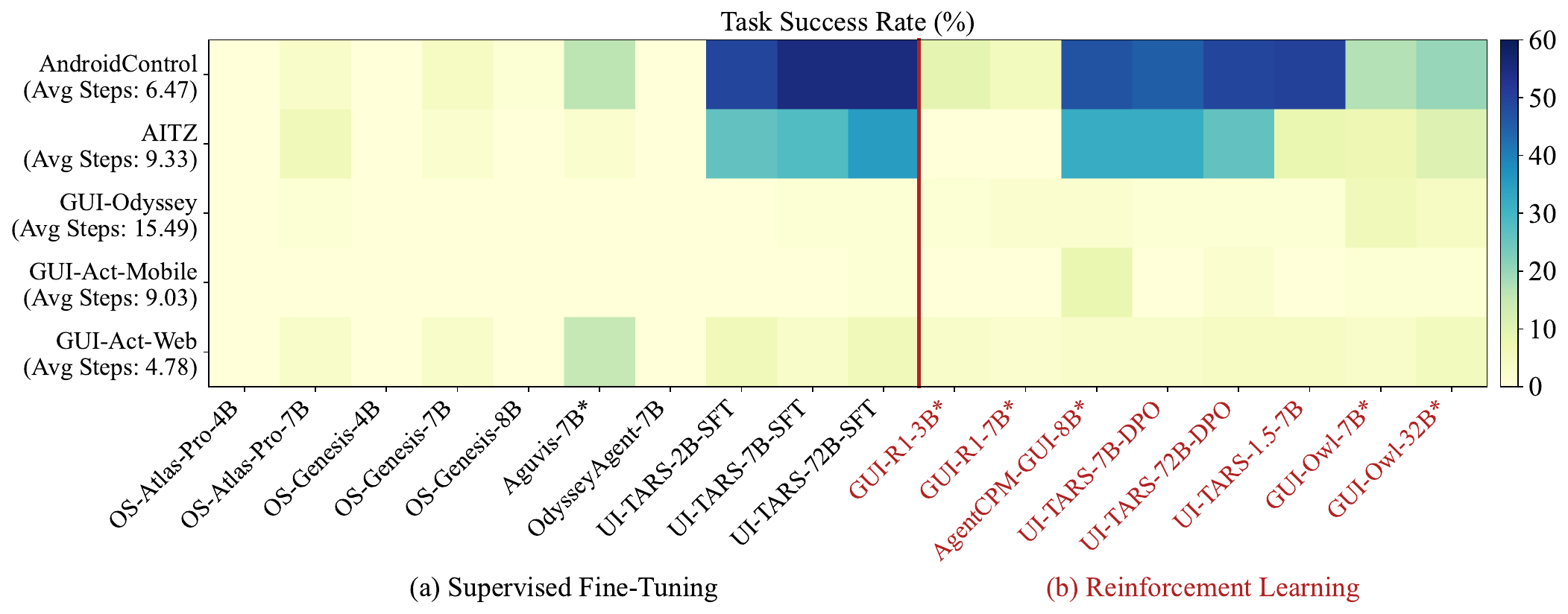}
    \caption{Task success rates for multimodal agents across five datasets. Models on the x-axis are grouped by training paradigm. The y-axis lists datasets, with parentheses indicating each dataset’s average interaction lengths (Avg Steps). ``*'' denotes models providing CoT for action reasoning.}
    \label{task success rate}
    \vspace{-0.3cm}
\end{figure}

As shown in Figure~\ref{task success rate}, we evaluate 18 models on 5 benchmark datasets. Our results show that early SFT-based agents consistently underperform, underscoring the limitations of relying on a single training paradigm. In contrast, later models achieve substantial gains as data coverage, model capacity, and training sophistication increase. The improvement is most evident on datasets such as AndroidControl and AITZ, which likely reflects exposure to similar scenarios during training. However, these gains do not generalize: performance drops markedly on long-horizon tasks (e.g., GUI Odyssey) and alternative platforms (e.g., GUI Act-Mobile, Web). This pattern highlights the intrinsic limits of current generalization. Although scaling and strategy optimization deliver clear benefits, agents remain tightly coupled to the distributions seen in training.

Building on these results, we further highlight two complementary findings. (i) Although RL is commonly regarded as a pathway to generalization, we observe that SFT-based models outperform their RL counterparts. (ii) CoT-augmented agents (e.g., AgentCPM-GUI) not only provide decision interpretability but also match the performance of the strongest non-reasoning models. In contrast to the conclusions of~\citep{zhang2025does}, our results suggest that CoT remains essential to advance multimodal agents, although still imperfect. To this end, we provide a quantitative analysis of memory and reasoning to explain the limited generalization of multimodal agents.

\section{Agent-ScanKit}
\begin{figure}[!t]
    \centering
    \includegraphics[width=1\linewidth]{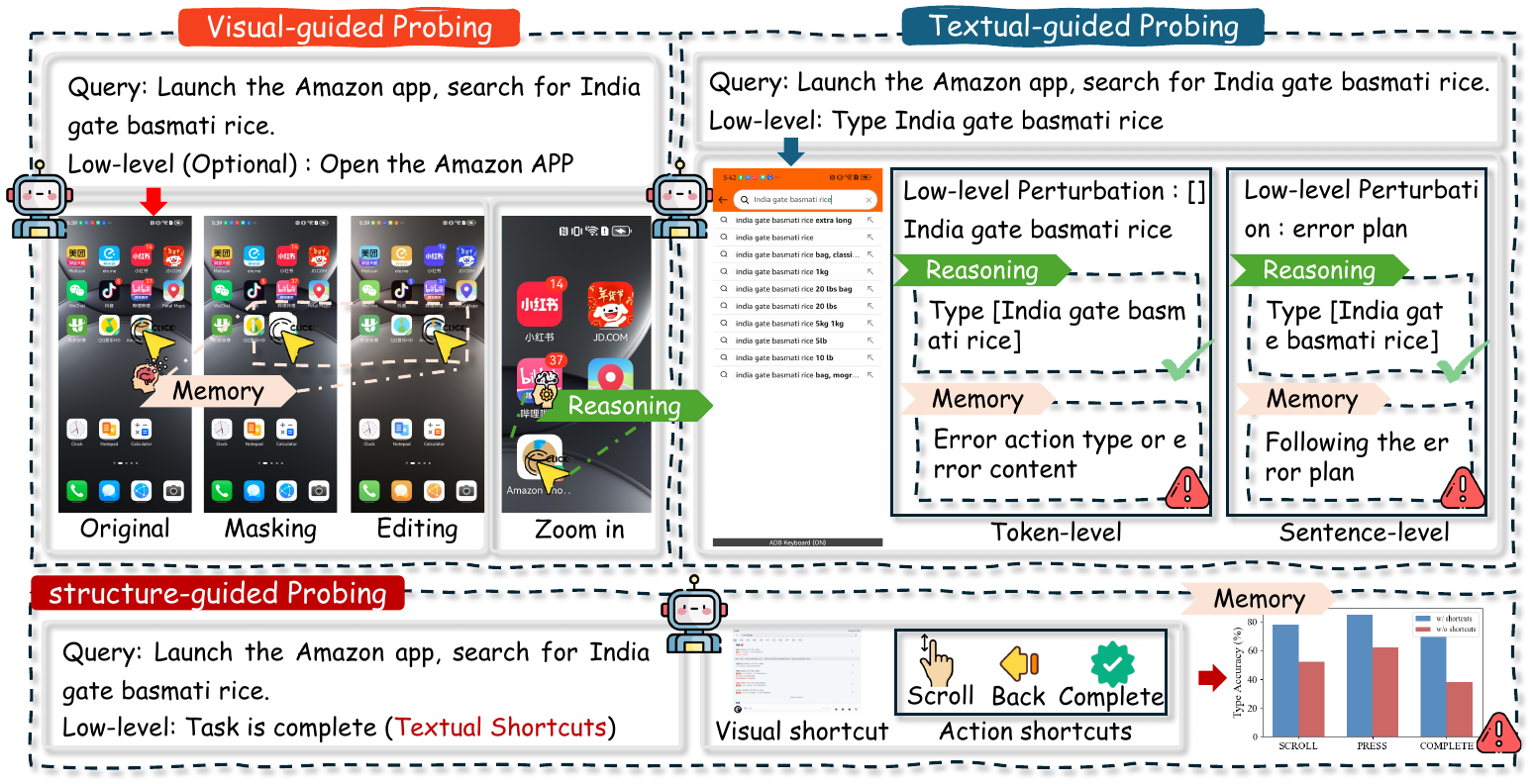}
    \caption{Overview of the Agent-ScanKit framework. The framework systematically probes multimodal agents with controlled perturbations along visual, textual, and structural dimensions, revealing the interplay between memory and genuine reasoning.}
    \label{pipeline}
    \vspace{-0.3cm}
\end{figure}

Based on the observations in Section~\ref{sec3}, we propose \textbf{Agent-ScanKit}, a probing toolkit that systematically quantifies the memory and reasoning capabilities of multimodal agents under controlled input perturbations. As illustrated in Figure~\ref{pipeline}, Agent-ScanKit incorporates three orthogonal probing paradigms: (i) visual-guided, (ii) text-guided, and (iii) structure-guided.

Following the POMDP formulation of GUI tasks, we extend the perceptual state space $\mathcal{S}$  with perturbation operators $\mathcal{P}$, forming a perturbed POMDP:
\begin{equation}
    \mathcal{M}_p = (G, \mathcal{S}, \mathcal{A}, \mathcal{T}, \mathcal{R}, \mathcal{H}, \mathcal{P}),
\end{equation}
where $\mathcal{P}: S\rightarrow S$ modifies the observed state $s_t$ in time step $t$. Given a goal $g\in G$, the agent receives perturbed observations $s'_t=\mathcal{P}(s_t)$, and selects action according to: 
\begin{equation}
    a_t \sim\pi(a_t|s'_t,g).
\end{equation}
By contrasting agent performance under perturbed versus unperturbed conditions, we quantify perturbation sensitivity as:
\begin{equation}
    \Delta_P = \mathbb{E}_{(g,s_t)}[Acc(\pi(s_t, g)) - Acc(\pi(\mathcal{P}(s_t), g))].
\end{equation}
Similarly, goal perturbations $\mathcal{P}: G \rightarrow G$ is used to probe text- and structure-guided mechanisms.

\subsection{Visual-guided Probing}
In the visual-guided level, we hypothetical existing agents often exploit positional priors (e.g., ``confirm buttons usually appear at the bottom-right'') rather than reasoning over screen content. To quantify such visual memory biases, we propose visual-guided perturbations: object masking and editing by obscuring or removing targets, while zoom-in introduces an OOD-like state to evaluate reasoning beyond rote memorization (Figure~\ref{pipeline}).

\noindent\textbf{Object Masking \& Editing.} We introduce two operators that directly alter the ground-truth target element $e^* \subseteq s_t$:
\begin{equation}
\mathcal{P}\left(s_t, e^*\right): \quad s_t^{\prime}(x, y)= \begin{cases}0, & (x, y) \in \Omega\left(e^*\right) \\ s_t(x, y), & \text { otherwise }\end{cases}
\end{equation}
where $\Omega\left(e^*\right)$ denotes the spatial region of $e^*$. Object masking can remove perceptual evidence of $e^*$, evaluating whether the agent relies on memorized spatial priors.

To probe the depth of spatial memory, we employ object editing, a perturbation that eliminates target elements and reconstructs them with interpolation over neighboring pixels:
\begin{equation}
\mathcal{P}\left(s_t, e^*\right): \quad s_t^{\prime}(x, y)= \begin{cases}\delta\left(s_t(x, y)\right), & (x, y) \in \Omega\left(e^*\right) \\ s_t(x, y), & \text { otherwise },\end{cases}
\end{equation}
where $\delta(\cdot)$ is the image-editing algorithm. If the performance remains stable under perturbations, i.e., when $\Delta_\mathcal{P}$ is small, it indicates that the agent's is primarily driven by memory retrieval rather than genuine reasoning, and vice versa.

\noindent\textbf{Zoom-in.} To evaluate reasoning, we define the zoom operator:
\begin{equation}
\mathcal{P}\left(s_t, q^*\right): \quad s_t^{\prime}=\operatorname{Crop}\left(s_t, q^*\right),
\end{equation}
where $\{q_1, q_2, q_3, q_4\}$ partition the UI into quadrants, and $q^*$ is the quadrant containing the target $e^*$. Notably, zoom-in removes global layout information while preserving local fidelity. If the agent successfully identifies $e^*$ under $\mathcal{P}$, i.e., $\Delta_\mathcal{P}$ is small, it demonstrates contextual reasoning, vice versa.

\subsection{Text-guided Probing} Multimodal agents operate in a joint visual–textual space, where the textual goal $g \in G$ guides perception and decision-making. Yet, agents remain under-optimized in navigating the vast lexical space. It is unclear whether this stems from memorizing atomic instructions or from an inability to reason over user queries. To probe this distinction, we introduce text-guided perturbations at both the token and sentence levels in the low-level setting.

\noindent\textbf{Token-level.} We hypothesize that starting words in the atomic instruction are pivotal for memory-driven behavior under the given text $g$. Accordingly, we modify the instruction as $g' = g \backslash\left\{w_i\right\}$. If $w_i$ is correctly inferred, $\Delta_\mathcal{P}$ should remain sufficiently small, and vice versa.

\noindent\textbf{Sentence-level.} We hypothesize that atomic instructions are central to memory-driven text processing. To evaluate it, we substitute the instructions by setting $g^{\prime} = \tilde{g}$ that $\tilde{g} \neq g$. If $g^{\prime}$ can be disregarded, $\Delta_\mathcal{P}$ should remain small, and vice versa.

\subsection{Structure-guided Probing}
As discussed in Section~\ref{sec3}, multimodal agents exhibit inherent optimization biases, which manifest as suboptimal reflection and state–action decisions. We conjecture that such behaviors arise from two systematic memory shortcuts internalized by the model: visual shortcuts and action shortcuts.

\noindent\textbf{Visual Shortcuts.} We define a visual shortcut as the model relying solely on the current screen $s_t$ for reflection or state-action decisions. Formally, if retaining only the current screen $s_t$ allows the agent to generate the action $a_t = \pi(a_t \mid s_t)$ with minimal $\Delta_\mathcal{P}$, then the agent's behavior is dominated by visual shortcuts; otherwise, no such shortcut exists.

\noindent\textbf{Action Shortcuts.} We define a action shortcut as the case where the model relies exclusively on the atomic instruction $g$ for reflection or state decision-making. Formally, if retaining only $g$ enables the agent to generate the action $a_t = \pi(a_t \mid g)$ with minimal $\Delta_\mathcal{P}$, then the agent’s behavior is dominated by action shortcuts; otherwise, no such shortcut exists.

\section{Experiments}
We first outline the experimental setup and then present the main results. More experimental details and results can be found in the Appendix~\ref{setup} and Appendix~\ref{more results}.
\subsection{EXPERIMENT SETUPS}\label{ssetup}
\noindent\textbf{MLLM-based GUI Agents.} We evaluate 18 representative open-source models developed by 8 institutions, spanning diverse architectural and training paradigms. These include the OS-Atlas series~\citep{wu2024atlas}, the OS-Genesis series~\citep{sun2024genesis}, the UI-TARS series~\citep{qin2025ui}, Aguvis-7B~\citep{xu2025aguvis}, OdysseyAgent-7B~\citep{lu2024gui}, the GUI-R1 series~\citep{luo2025gui}, the Mobile-Agent series~\citep{ye2025mobileagentv3foundamentalagentsgui}, and AgentCPM-GUI-8B~\citep{zhang2025agentcpmgui}. 

\noindent\textbf{Evaluation Benchmarks.} We evaluate five representative agent benchmarks across three different platforms: AndroidControl~\citep{li2024effects}, AITZ~\citep{zhang2024android}, GUI-Odyssey~\citep{lu2024gui}, and GUI-Act-Mobile~\citep{chen2024guicourse} for mobile agents; GUI-Act-Web and OmniAct-Web~\citep{kapoor2024omniact} for web agents; and OmniAct-Desktop for Windows environments. 

\noindent\textbf{Settings.} In our benchmark evaluation, we report model performance under high-level and, where applicable, low-level settings. For visual-guided probing, we evaluate on the grounded samples from Table~\ref{tab_a1}. For the text-guided probing, we use samples that require textual input (e.g., \textsc{Type}, \textsc{OpenApp}) from Table~\ref{tab_a1}. For structure-guided probing, we focus on reflective actions (\textsc{PressBack}, \textsc{PressHome}) and state actions (\textsc{Wait}, \textsc{Complete}). All evaluations employ the official open-source prompts and inference parameters. Unless otherwise specified, each agent is conducted under the low-level setting using samples with 100\% step-wise accuracy.

\noindent\textbf{Metrics.} We use four metrics to evaluate all multimodal agents: accuracy of action-type prediction (Type), accuracy of coordinate prediction (Grounding), step-wise success rate (SR), and task success rate (TSR). Unless otherwise specified, we report $\Delta P_{\text{Type}}$ and $\Delta P_{\text{SR}}$ in probing experiments. In addition, we introduce two complementary metrics in visual-guided probing: visual memory consistency (VMC) and reflection score (RS).

\subsection{Main Results and Analysis}
We present key findings at three levels of probing: visual-guided (Section~\ref{5.2.1}), text-guided (Section~\ref{5.2.2}) and structure-guided (Section~\ref{5.2.3}), each revealing a distinct balance between memory and reasoning in multimodal agents in GUI tasks.

\subsubsection{Analysis of Visual-Guided Level}\label{5.2.1}
\begin{table}[t]
    \centering
    \Huge
    \caption{Visual-Guided Probing on 7$\sim$8B multimodal agents in GUI tasks. Memory is evaluated through object masking and editing, whereas reasoning is assessed via zoom-in. The masking and editing ratio, and the distance threshold of VMC are set to 50 pixels.}
    \label{tab:7b8b}
    \renewcommand{\arraystretch}{1.02}
    \resizebox{\linewidth}{!}{
    \begin{tabular}{lcccccccccccc}
    \toprule
    \multirow{2}{*}{\textbf{GUI Agents}} 
    & \multicolumn{4}{c}{\textbf{Object Masking}} 
    & \multicolumn{4}{c}{\textbf{Object Editing}} 
    & \multicolumn{4}{c}{\textbf{Zoom-in}} \\
    \cmidrule(lr){2-5} \cmidrule(lr){6-9} \cmidrule(lr){10-13}
     & $\Delta P_{\text{Type}}\downarrow$ & $\Delta P_{\text{SR}}\downarrow$ & VMC$\uparrow$ & RS$\uparrow$ 
       & $\Delta P_{\text{Type}}\downarrow$ & $\Delta P_{\text{SR}}\downarrow$ & VMC$\uparrow$  & RS$\uparrow$  
       & $\Delta P_{\text{Type}}\downarrow$ & $\Delta P_{\text{SR}}\downarrow$ & VMC$\downarrow$ & RS$\downarrow$   \\
    \midrule
    \rowcolor{gray!15}
    \multicolumn{13}{l}{\textbf{Supervised Fine-Tuning}} \\
    OS-ATLAS-Pro-7B & 9.8 & 44.8 & 45.3 & 8.55 & 8.5 & 42.6 & 39.1 & 7.52 & 13.6 & 40.5 & 0.72 & 12.3 \\
    OS-Genesis-7B & 1.1 & 7.3 & 94.4 & 0.18 & 3.0 & 13.7 & 83.5 & 0.87 & 2.9 & 98.8 & 86.0 & 1.61 \\
    OS-Genesis-8B & 0.2 & 7.4 & 98.8 & 0.12 & 0.4 & 7.7 & 98.7 & 0.14 & 1.8 & 96.2 & 95.6 & 0.09 \\
    OdysseyAgent-7B & 1.9 & 37.9 & 57.6 & 1.45 & 1.5 & 33.8 & 63.6 & 0.96 & 3.9 & 62.0 & 5.69 & 1.92 \\
    Aguvis-7B & 0.1 & 13.3 & 99.5 & 0.02 & 0.5 & 1.8 & 97.2 & 0.03 & 5.4 & 47.0 & 0.65 & 1.64 \\
    UI-TARS-SFT-7B & 9.4 & 34.0 & 69.7 & 5.48 & 7.2 & 33.8 & 60.9 & 5.12 & 12.1 & 36.5 & 1.43 & 8.66 \\
    \rowcolor{gray!15}
    \multicolumn{13}{l}{\textbf{Reinforcement Learning}} \\
    GUI-R1-7B & 4.2 & 15.6 & 79.5 & 3.72 & 9.2 & 37.0 & 54.2 & 8.66 & 4.9 & 44.6 & 0.40 & 4.00 \\
    AgentCPM-GUI-8B & 19.0 & 49.8 & 46.0 & 18.2 & 13.2 & 36.4 & 56.0 & 12.6 & 14.2 & 42.9 & 1.31 & 13.7 \\
    UI-TARS-DPO-7B & 9.0 & 35.6 & 56.3 & 1.74 & 7.0 & 31.8 & 60.6 & 4.04 & 14.3 & 37.5 & 1.13 & 9.43 \\ 
    UI-TARS-1.5-7B & 25.6 & 44.9 & 60.6 & 4.04 & 4.3 & 27.2 & 64.7 & 2.55 & 23.7 & 56.6 & 0.77 & 2.06 \\
    GUI-Owl-7B & 15.3 & 43.5 & 49.0 & 13.7 & 13.5 & 41.0 & 54.8 & 12.2 & 8.6 & 38.9 & 0.36 & 7.23 \\
    \bottomrule
    \end{tabular}}
    \vspace{-0.4cm}
\end{table}

Given the limitations of multimodal agents in coordinate spaces of infinite scope, we present the results of visually-guided probing in Table~\ref{tab:7b8b}. Overall, our findings show that current multimodal agents rely heavily on tightly coupled spatial memory when performing GUI-based tasks,
\begin{wrapfigure}{r}{0.50\linewidth} 
    \centering
    \vspace{-0.3cm}
    \includegraphics[width=\linewidth]{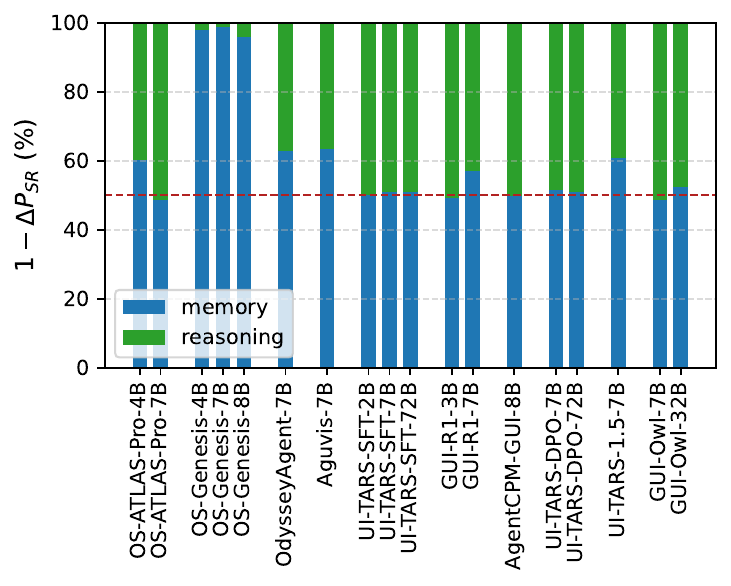}
    \vspace{-0.8cm}
    \caption{Distribution of memory and reasoning across multimodal agents of different scales.}
    \vspace{-0.4cm}
    \label{visual_model_scale}
\end{wrapfigure}
and once this alignment is perturbed, their reasoning capacity deteriorates sharply, leading to unstable behavior. 
In memory detection, agents fail to account for visual anomalies and instead resort to mechanical clicking, resulting in persistently low $\Delta P_{\text{Type}}$ and RS. Furthermore, $\Delta P_{\text{SR}}$ and VMC expose a strong bias toward selecting coordinates near the original predictions, highlighting the dependence on spatial memory. In reasoning probing, agents struggle to localize targets within the local context, resulting in elevated $\Delta P_{\text{SR}}$. In particular, OS-Genesis achieves $\Delta P_{\text{SR}}$ above 95\% and VMC exceeds 85\%, underscoring its heavy dependence on coordinate memorizing. In addition, we found that both $\Delta P_{\text{Type}}$ and RS still remain low, suggesting the presence of a potential text-oriented memory mechanism in agents. Compared to SFT, RL-based agents, particularly those that leverage explicit reasoning chains (e.g. AgentCPM-GUI and GUI-Owl), mitigate memory bias and exhibit stronger reflective capability. However, this reflexivity introduces notable side effects for reasoning. Once spatial memory is disrupted, they will generate over-reflection. 

In Figure 5, we illustrate the interplay between memory and reasoning via 1-$\Delta P_{\text{SR}}$. Early SFT models are clearly memory-dominated, but with larger datasets and improved training strategies, their reasoning ability strengthens, yet soon reaches a bottleneck. Meanwhile, memory effects persist and undermine reliability, as exemplified by UI-TARS-1.5 and GUI-R1. Moreover, scaling up models or incorporating RL does not consistently improve reasoning beyond their predecessors and may even introduce side effects, underscoring that larger parameter counts primarily expand spatial memory capacity rather than reasoning capability. The distribution of VMC and RS, together with ablation analysis, shows that agents' memory and reflection rely on the integrity of visual memory (Appendix~\ref{b3},~\ref{ablations}). Moreover, attention visualizations reveal their reasoning mechanisms while exposing the memory-driven nature of CoT (Appendix~\ref{vis}).

\subsubsection{Analysis of Text-Guided Level}\label{5.2.2}
Given the limitations of multimodal agents in the vocabulary space, we report text-guided probing results for input-dependent actions (e.g., \textsc{OpenApp}, \textsc{Type}) in Table~\ref{tab2}. 
\begin{wraptable}{r}{0.55\textwidth} 
    \centering
    \vspace{-0.6cm}
    \caption{Text-guided probing of multimodal agents with token-level and sentence-level evaluation.}
    \label{tab2}
    \renewcommand{\arraystretch}{1.1}
    \resizebox{\linewidth}{!}{
    \begin{tabular}{lcccc}
    \toprule
    \multirow{2}{*}{\textbf{GUI Agents}} 
    & \multicolumn{2}{c}{\textbf{Token-level}} 
    & \multicolumn{2}{c}{\textbf{Sentence-level}} \\
    \cmidrule(lr){2-3} \cmidrule(lr){4-5}
     & $\Delta P_{\text{Type}}\downarrow$ & $\Delta P_{\text{SR}}\downarrow$ 
     & $\Delta P_{\text{Type}}\uparrow$ & $\Delta P_{\text{SR}}\uparrow$  \\
    \midrule
    \rowcolor{gray!15}
    \multicolumn{5}{l}{\textbf{Supervised Fine-Tuning}} \\
    OS-ATLAS-Pro-7B & 3.90 & 14.8 & 9.50 & 20.1  \\
    OS-Genesis-7B & 30.5 & 57.1 & 67.7 & 85.4 \\
    Aguvis-7B & 0.30 & 5.20 & 3.10 & 6.60 \\
    UI-TARS-7B-SFT & 3.40 & 34.8 & 70.4 & 76.0 \\ \midrule
    \rowcolor{gray!15}
    \multicolumn{5}{l}{\textbf{Reinforcement Learning}} \\
    GUI-R1-7B-SFT & 5.30 & 16.8 & 20.9 & 33.2 \\
    AgentCPM-GUI-8B & 1.90 & 40.8 & 70.4 & 76.1 \\
    UI-TARS-DPO-7B & 4.70 & 18.8 & 50.5 & 63.3 \\
    UI-TARS-1.5-7B & 5.40 & 34.4 & 19.7 & 41.6 \\
    GUI-Owl-7B & 5.00 & 57.7 & 70.1 & 97.7 \\ \bottomrule
    \end{tabular}}
    \vspace{-0.4cm}
\end{wraptable}
At the token level, the omission of action start words does not affect the accuracy of action-type prediction, but it does reduce the accuracy of agents' vocabulary space predictions. At the sentence level, the UI-TARS series and RL-based agents show a stronger tendency toward instruction adherence, thereby avoiding mispredictions caused by overreliance on visual memory. Taken together, the two probing results indicate that lower values (e.g., OS-Atlas, Aguvis, UI-TARS-1.5, and GUI-R1) reflect dependence on visual memory. Although UI-TARS-SFT, UI-TARS-DPO, and Agent-CPM achieve higher semantic space prediction accuracy when adhering to instructions, their reliability remains limited. In other words, no agent can reliably infer input content from instructions lacking action-start words. 


\subsubsection{Analysis of Structural-Guided Level}\label{5.2.3}
\begin{table}[!t]
    \centering
    \caption{Structure-guided probing of multimodal agents on visual and action shortcuts. Higher values indicate stronger reliance; green = action, red = visual.}
    \label{tab3}
    \renewcommand{\arraystretch}{1}
    \resizebox{\linewidth}{!}{
    \begin{tabular}{lcccccccc}
    \toprule
    \multirow{2}{*}{\textbf{GUI Agents}} 
    & \multicolumn{4}{c}{\textbf{Visual Shortcuts}} 
    & \multicolumn{4}{c}{\textbf{Action Shortcuts}} \\
    \cmidrule(lr){2-5} \cmidrule(lr){6-9}
     & \textsc{Scroll} & \textsc{Wait} & \textsc{Press} & \textsc{Complete}
      & \textsc{Scroll} & \textsc{Wait} & \textsc{Press} & \textsc{Complete}  \\
    \midrule
    \rowcolor{gray!15}
    \multicolumn{9}{l}{\textbf{Supervised Fine-Tuning}} \\
    OS-ATLAS-Pro-7B & 71.3 & 67.2 & \cellcolor[HTML]{F4CCCC}
90.5 & 64.1 & 67.1 & \cellcolor[HTML]{C6EFCE}82.4 & 51.4 & 46.5 \\
    OS-Genesis-7B & 19.0 & 46.9 & 27.1 & 10.9 & 60.6 & 0.00 & \cellcolor[HTML]{C6EFCE}84.5& 33.6\\
    Aguvis-7B &49.8 &\cellcolor[HTML]{F4CCCC}87.5 &\cellcolor[HTML]{F4CCCC}99.6 &\cellcolor[HTML]{F4CCCC}89.2 & 47.8 & \cellcolor[HTML]{C6EFCE}96.8 & \cellcolor[HTML]{C6EFCE}94.8 & 0.13\\
    UI-TARS-7B-SFT &42.2 & 68.1 & 11.9 & 49.7 & 37.5 & \cellcolor[HTML]{C6EFCE}99.2 & \cellcolor[HTML]{C6EFCE}85.2 & \cellcolor[HTML]{C6EFCE}98.5  \\ \midrule
    \rowcolor{gray!15}
    \multicolumn{9}{l}{\textbf{Reinforcement Learning}} \\
    GUI-R1-7B-SFT & 28.9 & 69.2 &\cellcolor[HTML]{F4CCCC}97.8 & \cellcolor[HTML]{F4CCCC}88.0& 71.8 & 6.41 &  \cellcolor[HTML]{C6EFCE}80.2 & \cellcolor[HTML]{C6EFCE}96.9\\
    AgentCPM-GUI-8B &31.3 &\cellcolor[HTML]{F4CCCC}82.6 &5.43 &64.8 & 58.0 & \cellcolor[HTML]{C6EFCE}99.3 & \cellcolor[HTML]{C6EFCE}95.2 & \cellcolor[HTML]{C6EFCE}94.8 \\
    UI-TARS-DPO-7B & 30.5&75.3 &19.3& 43.1& 42.6 & \cellcolor[HTML]{C6EFCE}99.1 & \cellcolor[HTML]{C6EFCE}80.5 & \cellcolor[HTML]{C6EFCE}97.9\\
    UI-TARS-1.5-7B & 48.0 & 72.2 & 27.1 & 73.2 & \cellcolor[HTML]{C6EFCE}80.0 & \cellcolor[HTML]{C6EFCE}97.8 & \cellcolor[HTML]{C6EFCE}85.9 & \cellcolor[HTML]{C6EFCE}98.9 \\
    GUI-Owl-7B  &16.8 & 76.3 &1.83 & 75.6 &  17.9& 0.00& 2.23 & 74.3\\ \bottomrule
    \end{tabular}}
    \vspace{-0.3cm}
\end{table}
As discussed in Section 3, optimization for multimodal agents tends to emphasize coordinate and semantic aspects, leading to suboptimal performance on reflective and status actions. Table~\ref{tab3} quantifies the memory shortcuts induced by these actions in pursuit of training objectives. We observe that current models exhibit pronounced action shortcuts in \textsc{Wait} and \textsc{Press} actions, which require minimal visual involvement. Early SFT models such as OS-ATLAS and Aguvis show a stronger reliance on visual shortcuts, while state-of-the-art UI-TARS and RL models further amplify action shortcuts in the \textsc{Complete} action. Interestingly, GUI-R1 and AgentCPM also achieve high accuracy under visual shortcuts, highlighting a migration of reliance from action memory to visual memory. Moreover, \textsc{Scroll} results suggest that models exhibit reasoning through joint visual–semantic decision-making. Finally, GUI-Owl reflects a shift toward multimodal decision–reasoning, attributable to redesigned CoT and RL strategies.

\section{Conclusion}
We present Agent-ScanKit, a systematic probing toolkit for dissecting the memory and reasoning mechanisms of multimodal agents in GUI tasks. Our evaluation of 18 agents across five benchmarks revealed two core challenges: the infinite predictive space and finite generalization. Probing through three orthogonal paradigms further shows that these limitations arise from memory-dominated reasoning, leading to inherent unreliability. These results underscore that RL and CoT still require refinement to enhance the robust of multimodal agents in practical scenarios.


\section*{ETHICAL CONSIDERATIONS}
All authors of this work have read and agree to abide by the ICLR Code of Ethics. This work systematically investigates the causes of multimodal agent unreliability and their underlying reasoning mechanisms. All experiments were conducted in controlled environments using publicly available datasets and MLLMs. The results incorporated from prior work are licensed for standard research purposes and align with their intended use. In addition, we only used LLMs solely to aid with text polishing and language refinement. No LLM-generated content contributed to the conceptual development of this paper. Our three probing strategies are mutually orthogonal, each designed to analyze whether different action types are dominated by memory or by inference. Overall, this research is centered on advancing scientific understanding of multimodal agent robustness, with the aim of encouraging the community to develop more reliable decision-making mechanisms for multimodal agents. 

\section*{REPRODUCIBILITY STATEMENT}
We commit that all reported results are fully reproducible in this paper. The main text specifies our experimental setup (Section~\ref{ssetup}), with additional details provided in the Appendix~\ref{setup}. During the review stage, we provide supplementary materials including environment configurations, model download links, dataset preprocessing procedures, evaluation code for multimodal agents, and our sensitivity probing code. We also include sample evaluation logs to verify the authenticity of our results. We promise to release the complete codebase and preprocessing scripts to support transparency and community use.

\bibliography{iclr2026_conference}
\bibliographystyle{iclr2026_conference}

\clearpage

\appendix
\section{Detailed Experimental Setup}\label{setup}
\subsection{Action Space Mapping}\label{action_space}
The action space $\mathcal{A}$ is parameterized to capture common user interactions in GUI environments. We define $\mathcal{A}$ as a finite set of structured actions:
\begin{equation}
\begin{aligned}
\mathcal{A} = \big\{ 
\textsc{Click}(x,y),\ 
\textsc{Scroll}(d),\ 
\textsc{Type}(t),\ 
\textsc{PressBack},\ 
&\textsc{PressHome},\  
\textsc{Enter},\ \\
\textsc{Complete},\ 
\textsc{OpenApp},\ 
\textsc{Wait}\ 
\big\},
\end{aligned}
\end{equation}
where 
\begin{itemize}
    \item $\textsc{Click}(x,y)$ represents a click operation at normalized coordinates $(x,y) \in [0,1000]$ on the screen.  
    \item $\textsc{Scroll}(d)$ denotes a scroll action with discrete direction $d \in \{\text{up}, \text{down}, \text{right}, \text{left}\}$.  
    \item $\textsc{Type}(t)$ inputs a text string $t \in \mathcal{V}^*$, where $\mathcal{V}$ is the vocabulary.  
    \item $\textsc{PressBack}$ is to press the system \emph{back} button, typically used to return to the previous screen.   
    \item $\textsc{PressHome}$ is to press the system \emph{home} button, which minimizes the current application and returns to the device’s home screen.
    \item $\textsc{Enter}$ executes the \emph{enter} key, often confirming an input or submitting a form.  
    \item $\textsc{Complete}$ indicates the successful completion of the current task, signaling the termination of the interaction.  
    \item $\textsc{OpenApp}(t)$ launches a target application $t \in \mathcal{V}^*$ specified in the task context of AndroidControl benchmark.  
    \item $\textsc{Wait}$ pauses the agent’s execution for a predefined duration, useful in asynchronous or loading scenarios.  
\end{itemize}

This parameterization captures both spatially grounded actions (e.g., $\textsc{Clicks}$) and semantic actions (e.g., $\textsc{type}$ and $\textsc{scroll}$), enabling multimodal agents to operate in realistic software environments. It should be noted that the action space $\mathcal{A}$ exclusively selects shared actions, thus standardizing the evaluation criteria.

\subsection{Details of MLLM-based GUI Agents}\label{MLLM-based GUI Agents details}
Table~\ref{prompts} provides a systematic overview of representative multimodal agents in GUI domain, highlighting their foundation models, training paradigms, and reasoning capabilities. We observe that most agents leverage either the Qwen-VL or InternVL families, with a few adopting MiniCPM-based backbones. Training strategies vary between continued pretraining (CPT)~\citep{wu2024atlas}, supervised fine-tuning (SFT)~\citep{zhang2025agentcpmgui}, and reinforcement learning (RL)~\citep{tang2025magicgui}, reflecting the necessity of end-to-end performance improvement. In particular, only a small subset of agents incorporate RL-based optimization (e.g., DPO and GRPO), and the observed improvements remain limited in practice. The CoT column captures whether the model produces explicit reasoning traces. We also provide the availability of official prompt resources in the last column.
\begin{table}[t]
    \centering
    \caption{Overview of the evaluated multimodal GUI agents, including their foundation models and training paradigms. Here, CPT denotes continued pre-training, SFT denotes supervised fine-tuning, and RL denotes reinforcement learning. CoT indicates whether the model provides explicit reasoning processes. \textcolor{yellow}{\ding{52}\rotatebox[origin=c]{-9.2}{\kern-0.7em\ding{55}}} denotes models that output reasoning for high-level goals, while directly predicting actions for low-level goals. The final column reports the availability of official prompt resources.}
    \label{prompts}
    \renewcommand{\arraystretch}{1.4}
    \resizebox{\linewidth}{!}{
    \begin{tabular}{llccccc}
    \toprule
    \textbf{GUI Agents} & \textbf{Foundation Model} & \textbf{CPT} & \textbf{SFT} & \textbf{RL}  & \textbf{CoT} & \textbf{Prompt Links} \\ \midrule
     OS-Atlas-Pro-4B & InternVL-2-4B & \textcolor{green}{\ding{51}} &  \textcolor{green}{\ding{51}} & \textcolor{red}{\ding{55}} & \textcolor{red}{\ding{55}} & \multirow{5}{*}{\shortstack{\url{https://huggingface.co/}\\\url{OS-Copilot/}}}  \\ 
     OS-Atlas-Pro-7B & Qwen2-VL-7B & \textcolor{green}{\ding{51}} & \textcolor{green}{\ding{51}} & \textcolor{red}{\ding{55}} & \textcolor{red}{\ding{55}} &  \\ 
     OS-Genesis-4B & InternVL-2-4B & \textcolor{red}{\ding{55}} & \textcolor{green}{\ding{51}} & \textcolor{red}{\ding{55}} & \textcolor{red}{\ding{55}} & \\
     OS-Genesis-7B & Qwen2-VL-7B & \textcolor{red}{\ding{55}} & \textcolor{green}{\ding{51}} & \textcolor{red}{\ding{55}} & \textcolor{red}{\ding{55}} &  \\
     OS-Genesis-8B & InternVL-2-8B & \textcolor{red}{\ding{55}} & \textcolor{green}{\ding{51}} & \textcolor{red}{\ding{55}} & \textcolor{red}{\ding{55}} & \\ \midrule
     Aguvis-7B & Qwen2-VL-7B & \textcolor{green}{\ding{51}} & \textcolor{green}{\ding{51}} & \textcolor{red}{\ding{55}} & \textcolor{green}{\ding{51}} &\makecell[c]{\url{https://github.com/}\\\url{xlang-ai/aguvis}} \\ \midrule
     OdysseyAgent-7B & Qwen-VL-7B & \textcolor{red}{\ding{55}} & \textcolor{green}{\ding{51}} & \textcolor{red}{\ding{55}} & \textcolor{red}{\ding{55}} &
     \makecell[c]{\url{https://github.com/}\\\url{OpenGVLab/GUI-Odyssey/}} \\ \midrule
     UI-TARS-2B-SFT & Qwen2-VL-2B & \textcolor{green}{\ding{51}} &  \textcolor{green}{\ding{51}} &  \textcolor{red}{\ding{55}} & \textcolor{yellow}{\ding{52}\rotatebox[origin=c]{-9.2}{\kern-0.7em\ding{55}}} & \multirow{6}{*}{\shortstack{\url{https://github.com/}\\\url{bytedance/UI-TARS/blob/main/}\\\url{codes/ui_tars/prompt.py}}}\\ 
     UI-TARS-7B-SFT & Qwen2-VL-7B & \textcolor{green}{\ding{51}} &  \textcolor{green}{\ding{51}} &  \textcolor{red}{\ding{55}} & \textcolor{yellow}{\ding{52}\rotatebox[origin=c]{-9.2}{\kern-0.7em\ding{55}}} & \\ 
     UI-TARS-72B-SFT & Qwen2-VL-72B & \textcolor{green}{\ding{51}} &  \textcolor{green}{\ding{51}} &  \textcolor{red}{\ding{55}} & \textcolor{yellow}{\ding{52}\rotatebox[origin=c]{-9.2}{\kern-0.7em\ding{55}}} & \\ 
     UI-TARS-1.5-7B & Qwen2.5-VL-7B & \textcolor{green}{\ding{51}} &  \textcolor{green}{\ding{51}} & \textcolor{green}{\ding{51}} & \textcolor{green}{\ding{51}} & \\ 
     UI-TARS-7B-DPO & Qwen2-VL-7B & \textcolor{green}{\ding{51}} &  \textcolor{green}{\ding{51}} &  \textcolor{green}{\ding{51}} & \textcolor{yellow}{\ding{52}\rotatebox[origin=c]{-9.2}{\kern-0.7em\ding{55}}} & \\ 
     UI-TARS-72B-DPO & Qwen2-VL-72B & \textcolor{green}{\ding{51}} &  \textcolor{green}{\ding{51}} &  \textcolor{green}{\ding{51}} & \textcolor{yellow}{\ding{52}\rotatebox[origin=c]{-9.2}{\kern-0.7em\ding{55}}} & \\ \midrule
     GUI-R1-3B & Qwen2.5-VL-3B & \textcolor{red}{\ding{55}} & \textcolor{red}{\ding{55}} & \textcolor{green}{\ding{51}} & \textcolor{green}{\ding{51}} & \multirow{2}{*}{\makecell[c]{\url{https://github.com/}\\\url{ritzz-ai/GUI-R1}}} \\
     GUI-R1-7B & Qwen2.5-VL-7B & \textcolor{red}{\ding{55}} & \textcolor{red}{\ding{55}} & \textcolor{green}{\ding{51}} & \textcolor{green}{\ding{51}} &  \\ \midrule
     AgentCPM-GUI-8B & MiniCPM-V-8B & \textcolor{green}{\ding{51}} & \textcolor{green}{\ding{51}} & \textcolor{green}{\ding{51}} & \textcolor{green}{\ding{51}} & \makecell[c]{\url{https://huggingface.co/}\\\url{openbmb/AgentCPM-GUI}} \\ \midrule
     GUI-Owl-7B & Qwen2.5-VL-7B & \textcolor{green}{\ding{51}} & \textcolor{green}{\ding{51}} & \textcolor{green}{\ding{51}} & \textcolor{green}{\ding{51}} & \multirow{2}{*}{\makecell[c]{\url{https://github.com/}\\\url{X-PLUG/MobileAgent/tree}\\\url{/main/Mobile-Agent-v3/cookbook}}} \\ 
     GUI-Owl-32B & Qwen2.5-VL-32B & \textcolor{green}{\ding{51}} & \textcolor{green}{\ding{51}} & \textcolor{green}{\ding{51}} & \textcolor{green}{\ding{51}} & \\
    \bottomrule
    \end{tabular}}
\end{table}

\subsection{Details of Benchmarks}\label{datasets}
Table~\ref{dataset_statistics} provides a comprehensive overview of the benchmark datasets used in our evaluation, including the number of goals, screens, and the distribution of action types. This detailed characterization highlights the heterogeneity of interaction patterns across platforms, thereby evaluating the generalization of multimodal agents in task execution and environmental contexts.
\begin{table}[t]
    \centering
    \caption{Dataset statistics, including the number of goals, screens, and distribution over action types. ``–'' denotes that a dataset does not support a particular action type. Additionally, “Single” indicates datasets constructed for single-frame evaluation, while ``Multi'' refers to trajectory-level benchmarks that capture longer-horizon interactions.}
    \label{dataset_statistics}
    \Huge
    \renewcommand{\arraystretch}{1.4}
    \resizebox{\linewidth}{!}{
    \begin{tabular}{lccc|cccccccc}
     \toprule 
     \multirow{2}{*}{\textbf{Dataset}} & \multirow{2}{*}{\textbf{Type}}& \multirow{2}{*}{\textbf{Goal}} & \multirow{2}{*}{\textbf{Screen}} 
     & \multicolumn{8}{c}{\textbf{Action Space}} \\ \cmidrule(lr){5-12}
     & & & & \textsc{Click} & \textsc{Scroll} & \textsc{Type} &  \textsc{Press} & \textsc{OpenApp} & \textsc{Wait} & \textsc{Enter} & \textsc{Complete}  \\ \midrule
     AndroidControl & Multi& 1,543 & 9,987 & 5,083 & 1,211 & 632 &343 & 608 & 1,175 & - & 1543  \\
     AITZ & Multi& 506 & 4,724 & 2,736 & 601 & 500 & 265 & - &- & 118 & 506 \\
     GUI-Odyssey & Multi& 1,666 & 25,651 & 16,747 &2,622 & 2,666 & 2,044 & -& -& -& 1,572\\
     GUI-Act-Mobile& Multi & 230 &2,079 & 1,281 & 260 & 216 & - &-& - & 92 & 230  \\
     GUI-Act-Web & Multi & 66 & 316 & 97 & 149 & 26 & - & - & - &- & 44 \\
     GUI-Act-Web & Single & - & 1,410 & 1,089 & 211 & - & - & - & - & - & 110 \\
     OmniAct-Web & Single & - & 529 & 525 & - & - & - & - & - & - & - \\
     OmniAct-DeskTop & Single & - & 1,491 & 1,491 & - & - & - & - & -  & - & -\\
     \bottomrule
    \end{tabular}}
\end{table}

\subsection{Details of implementation}
Following~\citet{zhang2025agentcpmgui}, we evaluated 19 open source multimodal agents in five datasets using a unified benchmarking framework. Within Agent-ScanKit, for visual-guided probing, unless otherwise specified, we masked targets with a 50 black-pixel block, applied a 50-pixel edit during object modification, and in zoom-in tasks divided the screen into quadrants before selecting the target quadrant and magnifying it back to the original scale. For text-guided probing, token-level tasks replaced the initial word with \texttt{[]}, while sentence-level tasks injected the erroneous atomic instruction ``Click the Amazon APP''. For structure-guided probing, we corrupted the visual and textual modalities independently to identify the origin of memory shortcuts.

\subsection{Details of Evaluation Metrics}
For the standard metrics, Type denotes the exact match between the predicted and ground-truth action types (e.g., \textsc{Click} and \textsc{Scroll}). Grounding evaluates the accuracy of GUI grounding in downstream tasks. SR measures the step-level success rate, where a step is considered successful only if both the predicted action and its associated arguments (e.g., coordinates for a click action) are correct. For metrics of visual-guided probing, VMC denotes the difference between visual mask/editing and original, calculated as:
\begin{equation}
\text{VMC} = \frac{1}{N} \sum_{i=1}^{N} \mathbb{I} \left( \left\| \mathbf{p}_i^C - \mathbf{p}_i^O \right\|_2 \leq \gamma \right),
\end{equation}
where $\mathbf{p}_i^C \in \mathbb{R}^2$ and $\mathbf{p}_i^O \in \mathbb{R}^2$ denote the predicted coordinates in the original and masking/editing, respectively, for the $i$-th sample, and $\gamma$ is a distance threshold. $\mathbb{I}(\cdot)$ is the indicator function that returns 1 if the condition is true and 0 otherwise. RS indicates whether agents successfully trigger reflective actions (e.g., \textsc{PressBack} and \textsc{PressHome}) and status actions (e.g., \textsc{Complete} and \textsc{Wait}) when encountering masking/edited images.

In our evaluation, we report the SR for \textsc{Click}, \textsc{Type}, \textsc{OpenApp}, and \textsc{Scroll}. For \textsc{Scroll}, the direction argument (i.e., \textsc{Up}, \textsc{Down}, \textsc{Left}, and \textsc{Right}) must exactly match the ground truth. For \textsc{Type} and \textsc{OpenApp}, the predicted text and the ground truth must exactly match. For \textsc{Click}, following~\citet{zhang2024you}, we normalize predicted and ground-truth coordinates to 1000 and measure their relative distance. The prediction is considered correct if this distance is within 14\%. For other actions (e.g., \textsc{PressBack}), the prediction is considered correct only if it exactly matches the ground truth.

\section{More Results}\label{more results}
\subsection{Detailed Performance of multimodal GUI Agents across Action Types}\label{Action Types}

\begin{figure}[t]
    \centering
    \includegraphics[width=1\linewidth]{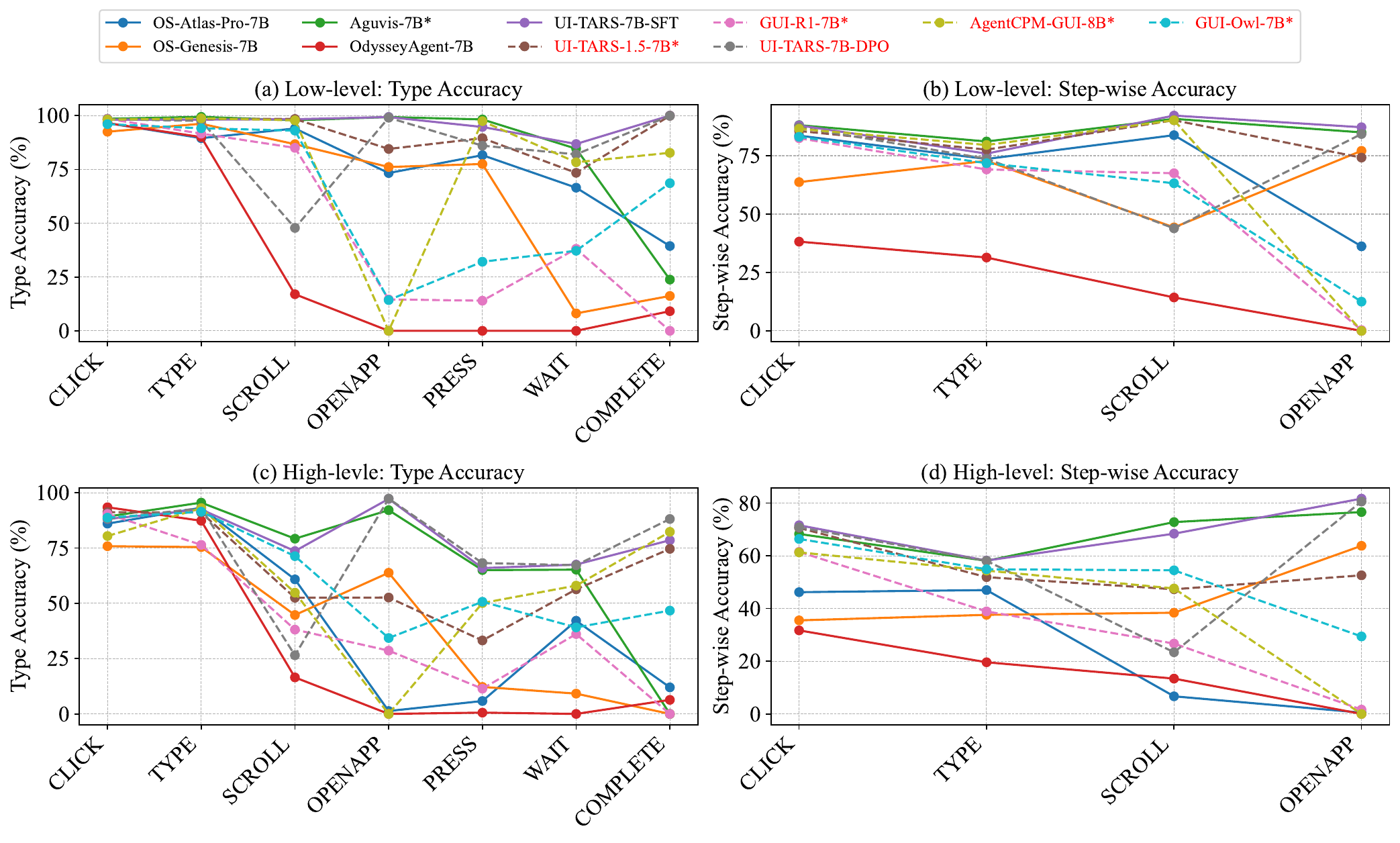}
    \caption{Detailed results of Type and SR between actions on AndroidControl Benchmark.}
    \label{细粒度指标2}
\end{figure}

As shown in Figure~\ref{细粒度指标2}, model performance is heavily focused on actions such as \textsc{Click} and \textsc{TYPE} in low- and high-level settings, while other action types exhibit varying degrees of instability. Introducing atomic instructions provides stronger textual guidance in both settings, improving accuracy. Similarly, RL-based models do not show a clear advantage over SFT counterparts. Finally, reasoning augmentation proves particularly helpful for high-level instructions, enabling models to reduce their reliance on explicit textual guidance.

\subsection{Detailed Performance of multimodal GUI Agents across Tasks and Platforms}\label{B.2}
Due to space constraints, Section~\ref{3.2} reports results only for 10 multi agents of 7$\sim$8B scale under the AndroidControl benchmark. For completeness, Tables~\ref{tab_a1} and~\ref{tab_a2} present the accuracy of 19 GUI agents ranging from 2B to 72B across four evaluation metrics. The results further corroborate the trends discussed in the main text. Early SFT models perform poorly across almost all benchmarks, highlighting the limitations of imitation-only training. In contrast, later SFT models benefit substantially from enhanced training strategies, larger datasets, and increased model scales, achieving consistent gains and in many cases outperforming RL-based agents.

A key factor driving this improvement is the reliance on atomic instructions (Low-level), which provide strong text-level guidance and significantly boost performance. This finding suggests that current multimodal agents behave more like single-step instruction followers than genuine reasoners. Notably, AgentCPM-GUI-8B, as a representative RL-based reasoning model, demonstrates clear advantages in high-level scenarios, validating the utility of CoT reasoning. However, even in this case, performance lags behind low-level settings that supply explicit textual guidance.

Despite these advances, generalization remains severely limited. Once extended to out-of-domain tasks or new environments, all models exhibit sharp performance degradation, underscoring that current GUI agents operate primarily under a memory-driven paradigm rather than robust reasoning-based generalization.
\begin{table*}[t]
\centering
\caption{Step-level and episode-level prediction performance on three GUI agent benchmarks, each containing both high-level and low-level instructions, is reported in terms of the success rates of Action Type, Grounding (Gr.), Step-wise Success Rate (SR), and Task Success. \textbf{Bold} and \underline{underlined} values denote the best and second-best results, respectively.}
\label{tab_a1}
\Huge
\renewcommand{\arraystretch}{1.4}
\resizebox{\linewidth}{!}{
\begin{tabular}{lcccccccccccc}
    \toprule
    \multirow{2}{*}{\textbf{GUI Agents}}  
    & \multicolumn{4}{c}{\textbf{AndroidControl-High/Low}} 
    & \multicolumn{4}{c}{\textbf{AITZ-High/Low}} 
    & \multicolumn{4}{c}{\textbf{GUI-Odyssey}} \\
    \cmidrule(lr){2-5} \cmidrule(lr){6-9} \cmidrule(lr){10-13} 
    & Type & Gr. & SR & TSR 
    & Type & Gr. & SR & TSR 
    & Type & Gr. & SR & TSR \\
    \midrule
    \rowcolor{gray!15}
    \multicolumn{13}{l}{\textbf{Supervised Fine-Tuning}} \\
    OS-Atlas-Pro-4B & 53.3/54.0 & 27.4/27.6 & 23.9/24.6 & 0/0 & 54.6/38.8 & 24.7/13.2 & 20.7/16.9 & 0/0 & 72.6/72.5 & 31.4/30.5 & 23.6/34.1 & 0/0\\    
    OS-Atlas-Pro-7B & 70.6/86.2 &61.2/83.6 &45.4/77.2 & 2/3 & 71.9/77.8 & 60.3/70.1 & 51.4/63.7 & 1/6 & \underline{90.1}/90.7 & 50.6/55.5 & 58.6/63.5 & 0/1 \\ \midrule
    OS-Genesis-4B & 42.6/69.9 & 24.4/61.1 & 16.7/45.0 & 0/0 & 30.4/65.2 &16.3/46.9 &11.1/45.4 & 0/0 & 26.2/46.6 & 0.53/1.06 & 5.49/9.17 & 0/0\\
    OS-Genesis-7B & 53.9/74.0 & 39.3/68.9 & 27.7/55.4 & 0/4 & 42.4/75.8 &31.5/55.7 & 21.8/53.9 & 0/2 & 24.0/53.8 & 0.64/8.38 & 3.19/19.1 & 0/0\\
    OS-Genesis-8B & 47.8/69.4 & 27.5/54.0 & 22.7/44.6 & 0/1 & 23.6/59.3 & 12.9/37.4 & 9.71/38.1 & 0/0 & 20.0/55.4 & 0.43/2.07 & 4.02/13.0 & 0/0\\ \midrule
    Aguvis-7B & 72.6/86.2 & 68.3/88.2 & 58.8/78.0 & 0/16 &  65.4/88.7& 53.4/80.2 & 44.7/76.0&  0/2& 81.1/81.2 & 55.5/55.7 & 59.8/59.9 & 0/0\\ \midrule 
    OdysseyAgent & 56.0/58.3  & 31.7/38.2 & 20.0/24.6 & 0/0 &  53.7/61.1 & 34.4/43.6 & 25.5/31.5 & 0/0 & 79.3/77.8 & \underline{71.3}/28.4 & 34.4/33.0 & 0/0 \\ \midrule
    UI-TARS-2B-SFT & 81.5/97.4 &66.7/86.1 & 67.7/\underline{87.6} & 17/49 & 76.5/98.9 & 61.5/85.1& 58.3/86.3 & 3/26 & 71.6/83.7 & 51.1/56.9 &45.3/55.0 &0/0\\
    UI-TARS-7B-SFT & 83.9/\textbf{97.7} & 71.9/87.8 &71.8/\textbf{89.6} & 22/\textbf{55} &  76.5/\underline{99.0} & 60.7/85.0 & 57.7/\underline{86.7} & 2/28 & 73.3/85.6 & 51.4/56.8 & 50.7/65.3 & 0/1\\
    UI-TARS-72B-SFT & \textbf{85.3}/\underline{97.5} & \underline{74.6}/\textbf{88.5} & \textbf{73.7}/\textbf{89.6}& \underline{23}/\textbf{55}& \underline{79.3}/\textbf{99.7} & 71.1/87.9 & \underline{63.7}/\textbf{88.8} & \underline{6}/\textbf{35} & 78.6/86.6 & 56.8/58.1 & 56.7/66.4 & 0/1\\
    \rowcolor{gray!15}
    \multicolumn{13}{l}{\textbf{Reinforcement Learning}} \\
    GUI-R1-3B &  60.0/77.0  & 48.5/73.7 & 38.6/62.3 & 2/9 &  53.5/79.0 & 37.9/74.1 & 26.5/56.8 & 0/0 & 67.6/86.4 & 40.4/61.6  & 35.0/62.3 & 0/1\\ 
    GUI-R1-7B & 64.1/75.0 & 61.5/82.5 & 44.4/62.7 & 3/5 & 55.9/84.1 & 41.2/78.5  & 28.4/57.2 & 0/0 & 73.1/\underline{91.1} & 43.8/66.6 & 37.1/61.7 & 0/2\\ \midrule
    AgentCPM-GUI-8B & 75.8/94.3 &61.3/86.5 &61.9/85.8 & 18/47 & \textbf{85.1}/95.5 & \textbf{74.6}/83.6 & \textbf{72.3}/86.2 & \textbf{16}/\underline{32} & \textbf{92.6}/\textbf{91.4} & 62.7/60.2 & \underline{67.8}/64.3 & 1/2 \\ \midrule
    UI-TARS-7B-DPO & 79.7/91.1 & 70.8/87.2 &  67.2/82.6& 22/45   & 77.5/97.4& 65.7/85.6 &57.4/\underline{86.7} & 2/\underline{32} & 71.9/86.5 & 53.9/61.0 & 49.7/61.6 & 0/1 \\
    UI-TARS-72B-DPO & \underline{84.0}/94.2 & \textbf{75.5}/\underline{88.4} & \underline{72.1}/86.6 & \textbf{24}/49 & 78.2/96.8 & \underline{74.3}/\textbf{88.2} & 61.9/86.0 & 5/26 & 76.5/84.3 & 58.2/61.2 & 52.6/60.5 & 0/1\\ 
    UI-TARS-1.5-7B &  78.2/96.0 & 70.6/87.5 & 64.1/\underline{87.6} & 15/\underline{50} & 76.4/88.1 & 66.2/85.6 & 56.5/77.6 & 3/18 & 78.8/88.3 &58.1/64.7 &51.3/64.5 & 0/1\\    \midrule
    
    GUI-Owl-7B & 72.8/80.7 & 66.4/83.1 & 56.9/69.0 & 9/17 & 76.7/85.1 & 59.5/69.3 & 56.7/70.0 & 2/7 & 81.4/84.9 & 68.1/\textbf{75.9} & 61.9/\textbf{70.7} & \underline{2}/\textbf{6} \\ 
    GUI-Owl-32B & 75.0/81.4 & 71.2/86.2 & 60.4/71.5 & 10/20 & 74.3/85.6 & 55.2/71.1 & 55.4/72.7 & 3/11 & 84.4/81.5 & \textbf{73.6}/\underline{75.0} & \textbf{68.9}/\underline{69.7} & \textbf{5}/\underline{4}\\ 
    \bottomrule
\end{tabular}}

\end{table*}

\begin{table*}[t]
\centering
\caption{Step-level and episode-level prediction performance across GUI Agent platforms, is reported in terms of the success rates of Action Type, Grounding (Gr.), Step-wise Success Rate (SR), and Task Success. \textbf{Bold} and \underline{underlined} values denote the best and second-best results, respectively.}
\label{tab_a2}
\renewcommand{\arraystretch}{1.3}
\resizebox{\linewidth}{!}{
\begin{tabular}{lcccccccccccccc}
    \toprule
    \multirow{2}{*}{\textbf{GUI Agents}}  
    & \multicolumn{4}{c}{\textbf{GUIAct-Mobile}} 
    & \multicolumn{4}{c}{\textbf{GUIAct-Web-Single/Multi}} 
    & \multicolumn{3}{c}{\textbf{Omniact-Desktop}} 
    & \multicolumn{3}{c}{\textbf{Omniact-Web}} \\
    \cmidrule(lr){2-5} \cmidrule(lr){6-9} \cmidrule(lr){10-12} \cmidrule(lr){13-15}
    & Type & Gr. & SR  & TSR 
    & Type & Gr. & SR  & TSR
    & Type & Gr. & SR  
    & Type & Gr. & SR  \\
    \midrule
    \rowcolor{gray!15}
    \multicolumn{15}{l}{\textbf{Supervised Fine-Tuning}} \\
    OS-Atlas-Pro-4B & 51.7 & 23.3&19.5 &0 & 51.4/45.2& 6.97/13.4 &9.50/14.2 &-/0&  73.2 & 15.7 & 15.6 & 39.5 & 0.95 & 0.94 \\ 
    OS-Atlas-Pro-7B & 61.7 &42.2 & 35.6 &0 & 88.9/53.8 & 81.2/16.5 & 75.0/27.5 & -/3 & 99.2& 80.6 &  80.4& 95.8 & 79.8 & 79.2 \\ \midrule
    OS-Genesis-4B &16.2 &1.40 &2.16&0& 78.4/31.0 & 58.5/13.4&54.8/9.17 &-/0 & 0.48 & 0.00 & 0.00 & 0.56 & 0.00 & 0.00\\
    OS-Genesis-7B & 24.9& 3.35&5.77 & 0& 84.8/32.6& 70.6/13.4 &64.7/13.0 &-/3 & 73.9 & 12.8&12.4 & 79.2 & 25.5 & 25.3  \\
    OS-Genesis-8B &11.9 & 1.32&1.53 &0& 65.1/32.9 &46.4/10.3 & 36.2/14.2 &-/0  &0.69 &0.00 & 0.00& 0.37 & 0.00 & 0.00\\ \midrule
    Aguvis-7B  & 50.5 & 33.0&28.6 &0 &82.6/50.6&81.1/46.4&71.2/40.8 &-/15 &90.6 &73.8 & 71.5& 93.0 & 78.6 & 78.1  \\ \midrule
    OdysseyAgent & 57.4& 18.0& 11.7& 0& 75.5/29.4& 3.94/2.06&3.04/1.26 &-/0 &95.1 &26.1 &26.0 &  90.9 & 26.8 & 26.6\\ \midrule
    UI-TARS-2B-SFT &\underline{72.9} &52.0 &42.8  & 0 & 81.7/57.3 & 64.4/38.1&61.6/47.5 &-/6 & 93.6 & 61.4 & 59.4 & 93.5 & 67.8 &67.2\\ 
    UI-TARS-7B-SFT & 19.2& 15.7&12.2 &0  & 86.8/45.2& 73.9/44.3& 70.6/33.2& -/3& 90.8 & 63.3&61.4 & 93.3 & 73.9 & 73.3\\ 
    UI-TARS-72B-SFT & 64.9&52.1 & 46.0 &1&  89.8/58.5& 72.4/54.6&  69.4/49.4 &-/6  &96.2 & 77.0&74.8 & 99.4 & 78.1 &78.1\\  \midrule
    \rowcolor{gray!15}
    \multicolumn{15}{l}{\textbf{Reinforcement Learning}} \\
    GUI-R1-3B & 48.4&16.3 &21.8 &0 & 43.4/27.2& 5.97/9.27& 7.16/10.5&-/3 & 86.0&77.6 &68.2 & 96.2 & 75.2 & 74.7\\
    GUI-R1-7B & 57.9& 27.3& 20.9 & 0& 69.4/35.4 &12.0/17.5&10.1/13.6 &-/2 & 85.0 & 79.6& 69.9 & 96.6 & 81.0 & 80.6 \\ \midrule
    AgentCPM-GUI-8B & \textbf{74.7} & \textbf{61.0} &\textbf{58.2} & \textbf{8} & 72.5/44.3& 40.9/14.4 &44.6/29.7 &-/3 &68.3 &44.7 &44.3 & 75.3 &46.8 &43.9  \\ \midrule
    UI-TARS-7B-DPO &53.6 &45.6 & 36.7 &0 & 87.6/50.0 &74.4/60.8 & 70.7/39.9 & -/3& 89.4 &64.1 & 61.9 & 96.9 & 70.6 & 70.1 \\ 
    UI-TARS-72B-DPO & 67.3 &\underline{53.3} & \underline{46.6} & \underline{2}& 86.9/38.3&73.0/56.7 &67.6/28.2  & -/4&85.4 & 75.3 &65.9 & 99.6 & 79.6 & 79.4\\
    UI-TARS-1.5-7B & 68.6&41.1 & 36.7& 0&  87.2/57.3& 70.6/44.3&67.1/42.7&-/4& 92.3& 51.5 & 49.8 & 98.5 & 84.8& 84.2 \\
    \midrule
    GUI-Owl-7B & 62.9 & 45.3 & 41.1 & 1 & 82.1/38.6 & 62.0/36.1 & 59.8/27.2 & -/3 &  88.6 & 65.4 & 65.2 & 91.7 & 71.5 & 70.4  \\
    GUI-Owl-32B & 60.9 &40.9 &38.3 & 1& 87.5/47.2 & 64.7/47.4 & 69.3/28.5 & -/5 & 92.5 & 71.1 &71.0 & 96.0 & 74.5 & 73.9  \\
    \bottomrule
\end{tabular}}
\end{table*}

\subsection{Further Results of Probing Experiments}\label{b3}
\subsubsection{Distribution of VMC and RS}
\begin{figure}
    \centering
    \includegraphics[width=1\linewidth]{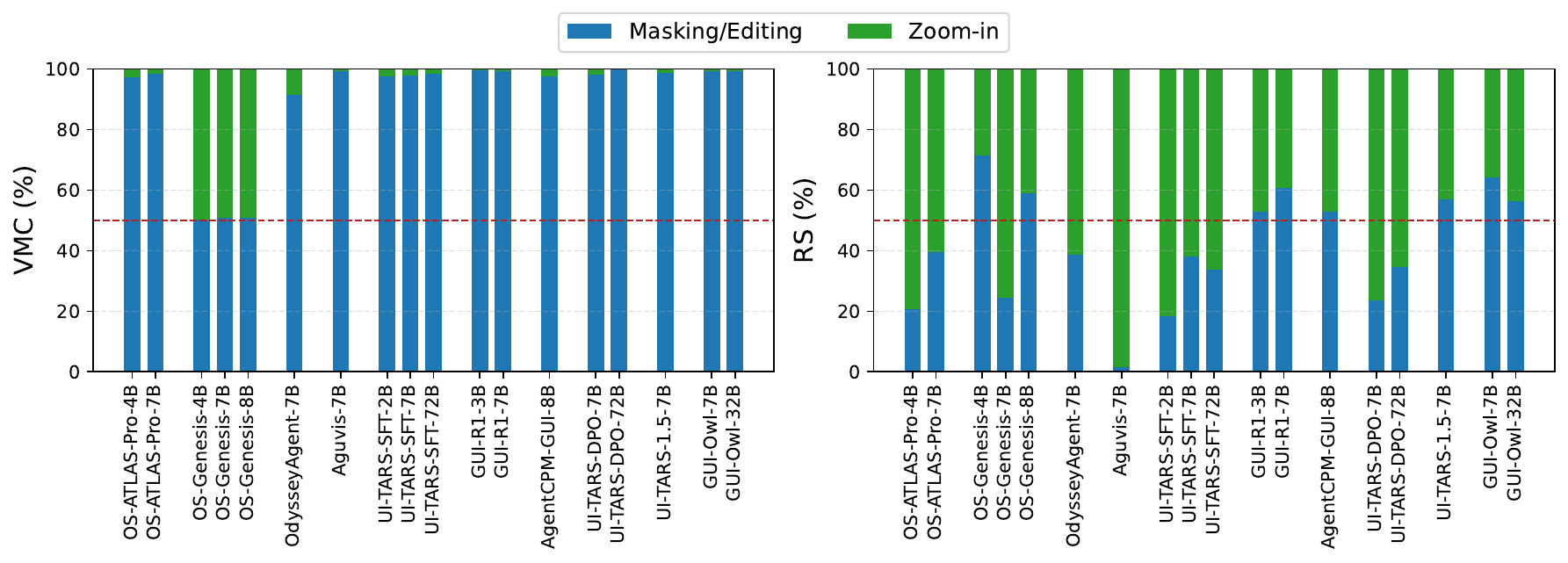}
    \caption{Distribution of VMC and RS in memory and reasoning across multimodal agents of varying model scales.}
    \label{vmc_vs_rs}
\end{figure}

Figure~\ref{vmc_vs_rs} illustrates the distribution of VMC and RS in memory and reasoning between multimodal agents on varying model scales. VMC directly reflects the agent's decision-making mechanism in memory and reasoning. When the ratio is evenly split, it indicates that the model relies heavily on memory, as seen in OS-Genesis. By contrast, when blue dominates, it suggests that the model activates memory when spatial memory is intact, but re-engages reasoning when spatial memory is disrupted. For RS, higher blue values and lower green values correspond to stronger reasoning ability. Early models almost universally exhibited over-reflection. With the expansion of training data and the refinement of training strategies, this effect was mitigated. However, models still tended to over-reflection when global visual information was impaired and underreflect when such information was preserved. These findings underscore that the agent’s reflection depends on the integrity of the global spatial context.

\subsubsection{Impact of Visual and Textual modalities on Visual-Guided Probing}\label{ablations}
\begin{figure}[t]
    \centering
    \includegraphics[width=0.8\linewidth]{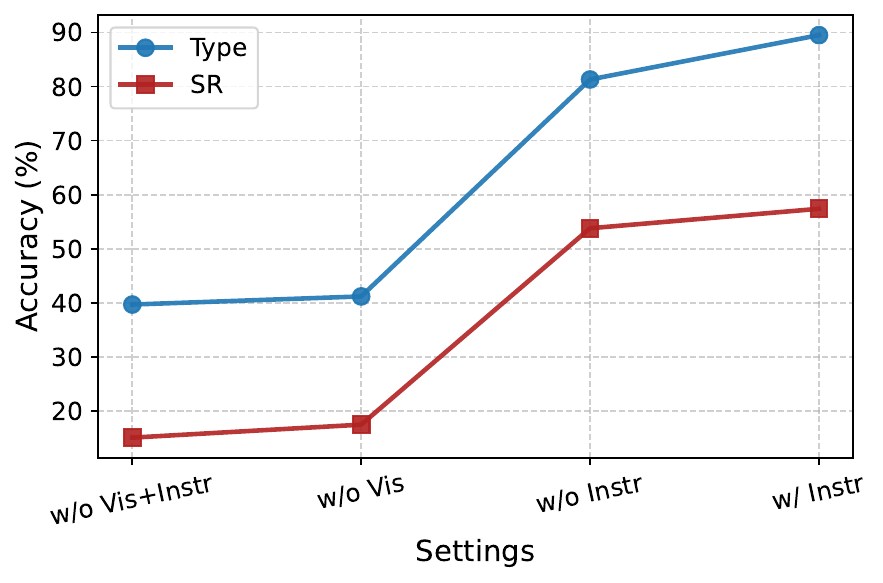}
    \caption{Ablation study of different settings in the object-masking of visual-guided probing. ``Vis'' = Visual Modality, ``Instr'' = Atomic Instruction.}
    \label{ablation}
\end{figure}

To further investigate the role of visual and textual modalities in grounding, we conduct an ablation analysis on object-masking probing under four conditions. As shown in Figure~\ref{ablation}, even without the visual modality and atomic instructions, UI-TARS-7B-SFT achieves 39.7\% Type accuracy and 15.1\% SR accuracy, revealing the contribution of absolute memory. Introducing the visual modality leads to a sharp increase to 81.3\% Type accuracy and 53.8\% SR accuracy, underscoring its importance for decision-making. However, this also indicates that the agent primarily activates global visual memory, resulting in a high rate of erroneous decisions. Notably, atomic instructions provide only a marginal benefit. When atomic instructions are removed, the agent's gains remained comparable to those achieved with instructions. This further supports the prediction that the agent relies on global visual memory rather than instruction-guided memorization.

\subsubsection{Visualization of Multimodal agents Attention}\label{vis}
\begin{figure}[t]
    \centering
    \includegraphics[width=1\linewidth]{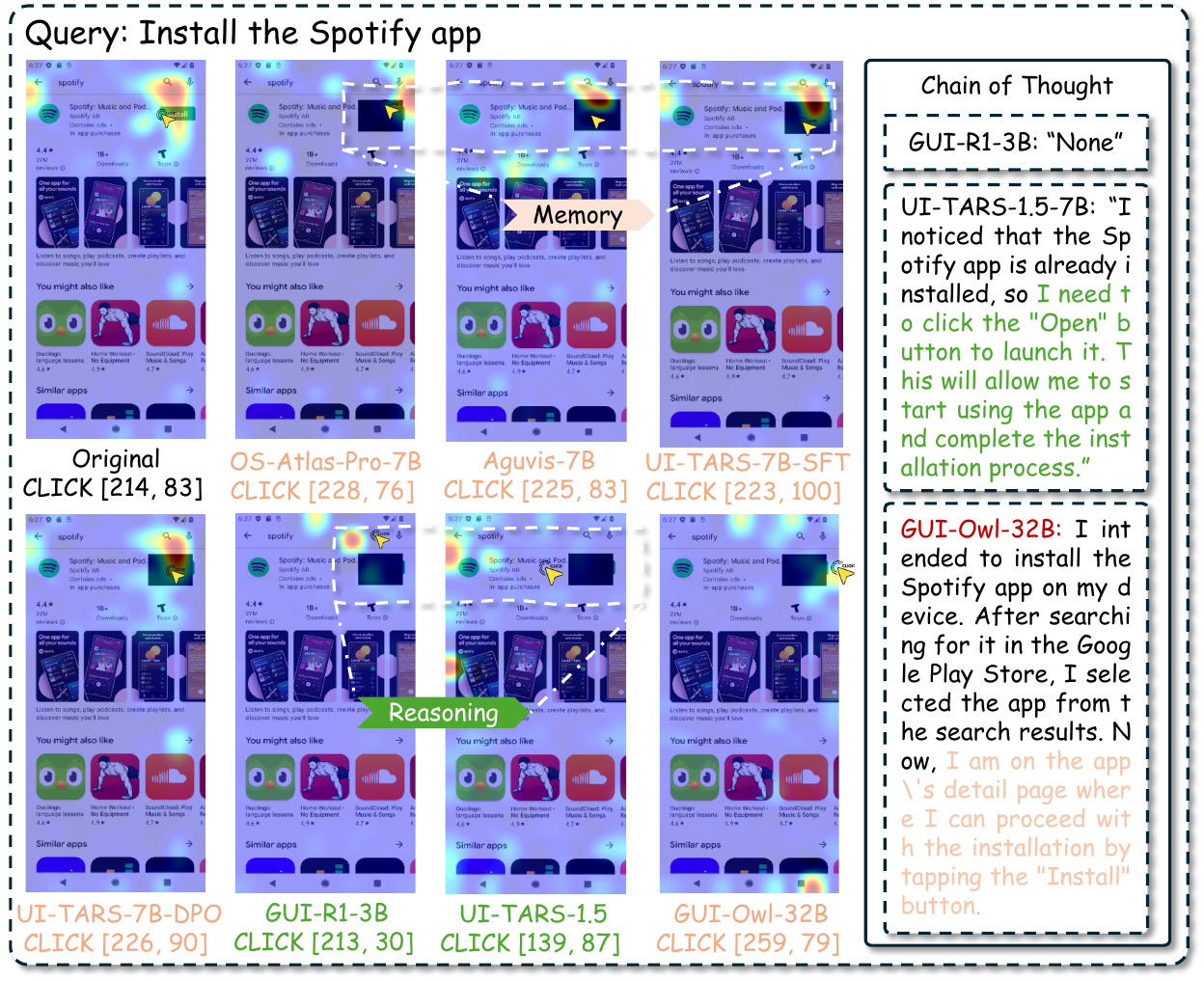}
    \caption{Visualization of relative attention in middle-layer during the decision-making process of Qwen-VL-based multimodal agents in the object masking probing. The right-side illustrate the reasoning trajectory with an integrated CoT agents.}
    \label{VMask}
\end{figure}

\begin{figure}[t]
    \centering
    \includegraphics[width=1\linewidth]{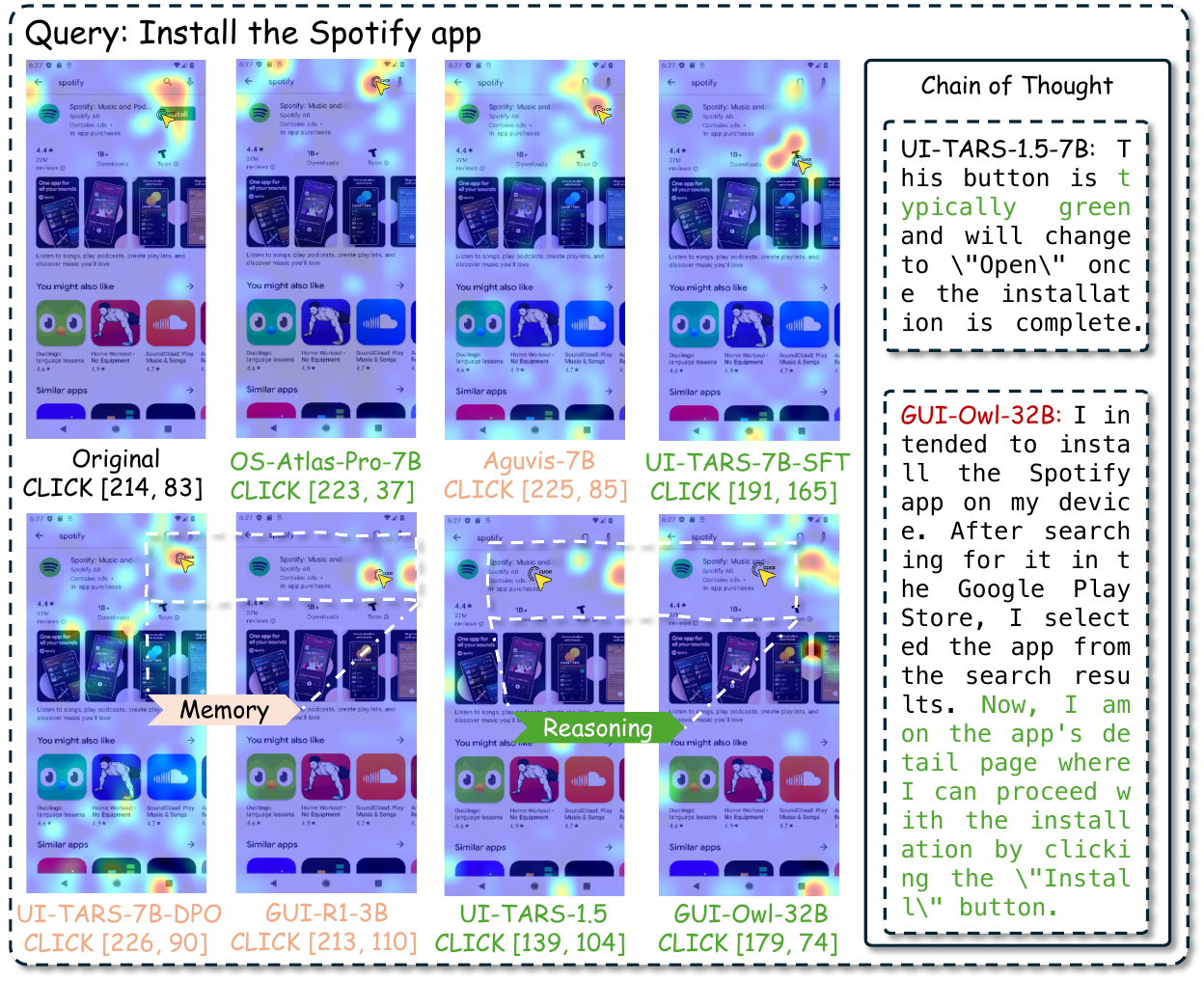}
    \caption{Visualization of relative attention in middle-layer during the decision-making process of Qwen-VL-based multimodal agents in the object editing probing. The right-side illustrate the reasoning trajectory with an integrated CoT agents.}
    \label{VEditing}
\end{figure}

Following~\citet{yan2025lasm} and~\citet{zhangmllms}, we adopt a relative attention-based visualization method to display the attention regions of multimoda agents. As shown in Figure~\ref{VMask},  the SFT model and UI-TARS-DPO preserve attention to object regions even under masking due to memory bias, thus generating coordinates consistent with the original. In contrast, GUI-R1 and UI-TARS-1.5 detect occlusions in the target areas, redirecting actions to the search box and the application details page, respectively. As shown in Figure~\ref{VEditing}, Aguvis and UI-TARS-DPO continue to exhibit memory-driven behavior in object editing, while OS-Atlas, UI-TARS-SFT, and UI-TARS-1.5 accomplish the task through exploratory strategies. Interestingly, GUI-Owl, equipped with CoT analysis, instinctively taps the target area under masking but switches to exploration during editing. We attribute this inconsistency to mechanical CoT generation. This enables memory recall rather than adaptive reasoning.

\end{document}